\documentclass{article}

    \PassOptionsToPackage{numbers, compress}{natbib}
\usepackage[preprint]{neurips_2020}

\usepackage[utf8]{inputenc} % allow utf-8 input
\usepackage[T1]{fontenc}    % use 8-bit T1 fonts
\usepackage{hyperref}       % hyperlinks
\usepackage{url}            % simple URL typesetting
\usepackage{booktabs}       % professional-quality tables
\usepackage{amsfonts}       % blackboard math symbols
\usepackage{amsmath}
\usepackage{amssymb}
\usepackage{nicefrac}       % compact symbols for 1/2, etc.
\usepackage{microtype}      % microtypography
\usepackage{graphicx}
\usepackage{xcolor}
\usepackage{bbm}
\usepackage{xspace}
\RequirePackage{authblk}

\newcommand{\ignore}[1]{}

\makeatletter
\DeclareRobustCommand\onedot{\futurelet\@let@token\@onedot}
\def\@onedot{\ifx\@let@token.\else.\null\fi\xspace}

\def\eg{\emph{e.g}\onedot}

\makeatother

\definecolor{MyDarkBlue}{rgb}{0,0.08,1}
\definecolor{MyDarkGreen}{rgb}{0.02,0.6,0.02}
\definecolor{MyDarkRed}{rgb}{0.8,0.02,0.02}
\definecolor{MyDarkOrange}{rgb}{0.40,0.2,0.02}
\definecolor{MyPurple}{RGB}{111,0,255}
\definecolor{MyRed}{rgb}{1.0,0.0,0.0}
\definecolor{MyGold}{rgb}{0.75,0.6,0.12}
\definecolor{MyDarkgray}{rgb}{0.66, 0.66, 0.66}

\usepackage[font={small}]{caption}
\usepackage{titlesec}
\titlespacing*{\section}{0pt}{0pt plus 2pt minus 2pt}{0pt plus 0pt minus 4pt}
\titlespacing*{\subsection}{0pt}{0pt plus 0pt minus 2pt}{-2pt plus 0pt minus 4pt}

\newcommand{\R}{\mathbb{R}}
\newcommand{\G}{\mathcal{G}}
\newcommand{\Th}{\Theta}
\renewcommand{\th}{\theta}

\newcommand{\PSGN}{\mathcal{F}^{\textbf{PSG}}}
\newcommand{\Rndr}{\mathcal{R}}
\newcommand{\GC}{\mathcal{GC}}
\newcommand{\F}{\mathcal{F}}
\newcommand{\ep}{\epsilon}

\newcommand{\D}{D}

\newcommand{\Seg}{\textbf{Seg}}
\newcommand{\B}{\mathcal{B}}
\newcommand{\SC}{\mathcal{SC}}
\newcommand{\np}{n_{\text{unroll}}}
\newcommand{\agg}{\mathbf{agg}}
\newcommand{\new}{\mathbf{new}}

\newcommand{\id}{\mathbbm{1}}

\renewcommand{\d}{\delta}

\newcommand{\dd}{\delta_{\text{dist}}}

\newcommand{\T}{\mathcal{T}}
\newcommand{\QTR}{\mathbf{QTR}}
\newcommand{\QSR}{\mathbf{QSR}}

\renewcommand{\a}{\mathbf{a}}
\newcommand{\SMCE}{\textbf{SoftMaxCE}}

\newcommand{\loss}{\mathcal{L}}
\newcommand{\proj}{\textbf{proj}}
\newcommand{\Prim}{\textbf{Primitives}}
\newcommand{\Play}{\textbf{Playroom}}
\newcommand{\Gib}{\textbf{Gibson}}
\newcommand{\vv}{\mathbf{v}}
\newcommand{\ww}{\mathbf{w}}

\title{Learning Physical Graph Representations from Visual Scenes}

\author[1,3,$\dagger$]{Daniel M. Bear}
\author[1,2]{Chaofei Fan}
\author[2]{Damian Mrowca}
\author[4]{Yunzhu Li}
\author[5]{Seth Alter}
\author[6]{Aran Nayebi}
\author[5]{Jeremy Schwartz}
\author[2,1]{Li Fei-Fei}
\author[2]{Jiajun Wu}
\author[5,4]{Joshua B. Tenenbaum}
\author[1,2,3]{Daniel L.K. Yamins}

\affil[1]{Department of Psychology, Stanford University}
\affil[2]{Department of Computer Science, Stanford University}
\affil[3]{Wu Tsai Neurosciences Institute, Stanford University}
\affil[4]{MIT CSAIL}
\affil[5]{MIT Brain and Cognitive Sciences}
\affil[6]{Neurosciences Ph.D. Program, Stanford University}
\affil[$\dagger$]{Correspondence: dbear@stanford.edu}

\begin{document}

\maketitle

\begin{abstract}

Convolutional Neural Networks (CNNs) have proved exceptional at learning representations for visual object categorization. 
However, CNNs do not explicitly encode objects, parts, and their physical properties, which has limited CNNs' success on tasks that require structured understanding of visual scenes.
To overcome these limitations, we introduce the idea of ``Physical Scene Graphs'' (PSGs), which represent scenes as hierarchical graphs, with nodes in the hierarchy corresponding intuitively to object parts at different scales, and edges to physical connections between parts.  
Bound to each node is a vector of latent attributes that intuitively represent object properties such as surface shape and texture. 
We also describe PSGNet, a network architecture that learns to extract PSGs by reconstructing scenes through a PSG-structured bottleneck.
PSGNet augments standard CNNs by including: recurrent feedback connections to combine low and high-level image information; graph pooling and vectorization operations that convert spatially-uniform feature maps into object-centric graph structures; and perceptual grouping principles to encourage the identification of meaningful scene elements.
We show that PSGNet outperforms alternative self-supervised scene representation algorithms at scene segmentation tasks, especially on complex real-world images, and generalizes well to unseen object types and scene arrangements. 
PSGNet is also able learn from physical motion, enhancing scene estimates even for static images. We present a series of ablation studies illustrating the importance of each component of the PSGNet architecture, analyses showing that learned latent attributes capture intuitive scene properties, and illustrate the use of PSGs for compositional scene inference.

%   The abstract paragraph should be indented \nicefrac{1}{2}~inch (3~picas) on
%   both the left- and right-hand margins. Use 10~point type, with a vertical
%   spacing (leading) of 11~points.  The word \textbf{Abstract} must be centered,
%   bold, and in point size 12. Two line spaces precede the abstract. The abstract
%   must be limited to one paragraph.
\end{abstract}

\section{Introduction}
\begin{figure}[t]
\centering
    \includegraphics[width=1.0\textwidth]{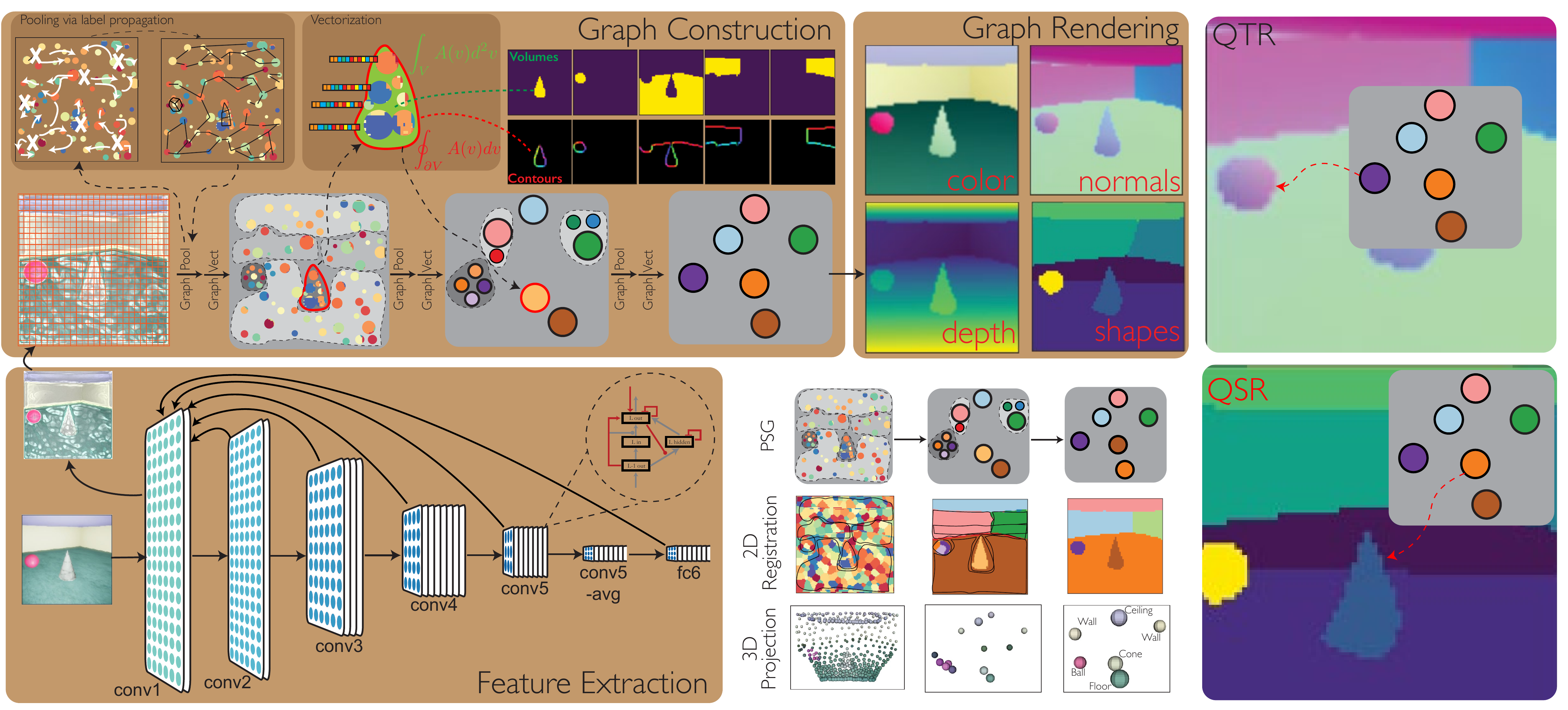}
\vspace{-15pt}
\caption{Overview of the Physical Scene Graph (PSG) representation and its construction and decoding by PSGNet. Brown boxes indicate the three stages of PSGNet: (1) Feature Extraction from visual input with a ConvRNN, (2) Graph Construction from ConvRNN features, and (3) Graph Rendering for end-to-end training. Graph Construction consists of a pair of learnable modules, Graph Pooling and Graph Vectorization, that together produce a new, higher PSG level from an existing one. The former dynamically builds a partition of pooling kernels over the existing graph nodes as a function of learned, pairwise node affinities (thresholded to become within-level graph edges, top left); the latter aggregates node statistics over the image regions (and their boundaries) associated with each pooling kernel to produce attribute vectors for the new nodes. Details of each stage and module are given in the Supplement. Three levels of an example PSG are shown (center, bottom) along with its quadratic texture (QTR) and shape (QSR) rendering (right.)}
\label{fig:model}
\vspace{-10pt}
\end{figure}

%CNNs are good at categorization but humans do more than categorization --
%really want object centric geometrically explicit solutions
To make sense of their visual environment, intelligent agents must construct an internal representation of the complex visual scenes in which they operate. 
Recently, one class of computer vision algorithms -- Convolutional Neural Networks (CNNs) -- has shown an impressive ability to extract useful categorical information from visual scenes.
However, human perception (and the aim of computer vision) is not only about image classification.  
Humans also group scenes into object-centric representations in which information about objects, their constituent parts, positions, poses, 3D geometric and physical material properties, and their relationships to other objects, are explicitly available.  
Such object-centric, geometrically-rich representations natively build in key cognitive concepts such as object permanence, and naturally support high-level visual planning and inference tasks.

%partially successful recent attempts
%Monet/IODINE bad b/c not good architecture
%MeshRCNN bad b/c supervised
Recent work, such as MONet~\cite{monet} and IODINE~\cite{iodine}, has made initial progress in attempting to equip CNNs with additional structures that aid in object-centric scene understanding. 
Learning from self-supervision with visual data, these works achieve some success at inferring latent representations for simple synthetic scenes.
However, these approaches are limited by architectural choices (\eg, the assignment of a fixed number of object slots per scene), and learn just from static images.  
As a result, they do not generalize effectively to complex real-world image data.
Recent approaches from 3D computer vision, such as 3D-RelNet~\cite{kulkarni20193d}, have attacked the question from a different point of view, combining key geometric structures (such as meshes) with more standard convolutional features. 
Such works achieve effective results, but do not offer a solution when detailed supervision is unavailable. 

%PSG representation abstractly
In this work, we propose a new representation, which we term Physical Scene Graphs (PSGs). 
The PSG concept generalizes ideas from both the MONet/IODINE and the 3D-RelNet lines of work, seeking to be at the same time geometrically rich enough to handle complex object shapes and textures, explicit enough about objects to support the top-down inferential reasoning capacities of generative models, flexible enough to learn from real-world or realistic (not toy) visual data, and self-supervisable, so as to not require extensive and laborious scene component labeling.

%PSG representation more concretely
PSGs represent scenes as hierarchical graphs, with nodes toward the top of the PSG hierarchy intuitively corresponding to larger groupings (\eg, whole objects), those nearer the bottom corresponding more closely to the subparts of the object, and edges representing within-object ``bonds'' that hold the parts together. 
PSGs are spatially registered, with each node tied to a set of locations in the convolutional feature map (and therefore image) from which it is derived.  
Node attributes represent physically-meaningful properties of the object parts relevant to each level of the hierarchy, capturing geometric aggregates such as object position, surface normals, and shape, as well as visual appearance.

%PSGNet overall
Our key contribution is a self-supervised neural network architecture, PSGNet, that learns to estimate PSGs from visual inputs. 
PSGNet involves several augmentations of the standard CNN architecture.  
To efficiently combine high- and low-level visual information during initial feature extraction, we add local recurrent and long-range feedback connections on top of a CNN backbone, producing a convolutional recurrent neural network (ConvRNN). 
We introduce a learnable Graph Pooling operation that transforms spatially-uniform ConvRNN feature map input into object-centric graph output.
We also introduce a novel way to summarize the properties of spatially-extended objects using a Graph Vectorization operation.  
These two operations are alternated in series to form a hierarchical graph constructor. 
To enforce that the graph constructor learns to identify intuitively meaningful nodes, we encode key cognitive principles of perceptual grouping into the model \cite{Spelke1990}, including both static and motion-based grouping primitives.
To ensure that learned node attributes contain disentangled geometric and visual properties of the latent object entities, we employ a novel graph renderer that decodes PSGs into feature maps. 

%results summary
On several datasets, PSGNet substantially outperforms alternative unsupervised approaches to scene description at segmentation tasks -- especially for real-world images. 
We also find that PSGNet usefully exploits object motion both when it is explicitly available and in learning to understand static images, leading to large improvements in scene segmentation compared to an alternative motion-aware self-supervised method. 
We find that the representation learned by PSGNet transfers well to new objects and scenes, suggesting that the strong constraints in the PSG structure force the system to learn key generalizable features even from limited training data. 
Analyzing the learned system, we show how the latent PSG structure identifies key geometric and object-centric properties, making it possible to compositionally "edit" the objects and attributes of inferred scenes. 
% Analyzing the learned system, we show how the latent PSG structure identifies key geometric and object-centric properties, including the proper handling of complex occlusions, making it useful for compositional scene inference tasks like scene editing. 

\textbf{Related Work.}
%\paragraph{Unsupervised segmentation from a single image.}
Unsupervised learning of scene representations is a long-standing problem in computer vision. Classic approaches study how to group pixels into patches and object instances~\cite{shi2000normalized, felzenszwalb2004efficient, isola2015learning}. Recently, machine learning researchers have developed deep generative models for unsupervised segmentation by capturing relations between scene components~\cite{Eslami2016Attend, monet, iodine, engelcke2019genesis}, with varying degrees of success and generality. 
Beyond static images, researchers have concurrently developed models that leverage motion cues for object instance segmentation from complementary angles~\cite{greff2017neural, kosiorek2018sequential}, aiming to jointly model object appearance and motion, all without supervision. These models are ambitious but only work on simple, mostly synthetic, scenes. Computer vision algorithms for motion clustering and segmentation from videos achieve impressive results~\cite{xie2019object,dave2019towards}, though they require high-quality annotations on object segments over time. A few papers~\cite{faktor2014video} have studied unsupervised segmentation or grouping from videos, which were later used to facilitate feature learning and instance segmentation~\cite{pathak2017learning,pathak2018learning}.
% However, these models have not yet shown an ability to generalize across different physical scenarios, likely because they do not explicitly represent objects and their physical properties.
However, these models do not employ an object-centric representation or model relations between scene elements.
Recent graph neural network models have shown promise at physical dynamics prediction, but require graph-structured input or supervision \cite{mrowca2018flexible, li2018learningParticle, li2020visual, sanchez2020learning} -- further motivating the unsupervised learning of graph representations from visual input.

Researchers have developed scene representations that take object-part hierarchy and inter-object relations into account. In particular, scene graphs~\cite{krishna2017visual} are a representation that simultaneously captures objects and their relations within a single image. While most related works on inferring scene graph representations from visual data require fine-grained annotations, several works have studied inferring objects and their relations from videos without supervision~\cite{steenkiste2018relational, stanic2019rsqair, xu2019unsupervised}. These models have the most similar setup as ours, but all focus on 2D pixel-level segmentation only. 
%Several works~\cite{mrowca2018flexible, li2019learning} also explore a hierarchical representation for objects and their parts in 3D, but assuming perfect perception to focus on learning object-centric dynamics prediction.
Several recent papers have explored building 3D, object-centric scene representations from raw visual observations. The most notable approaches include Factor3D~\cite{tulsiani2018factoring}, MeshRCNN~\cite{gkioxari2019mesh}, and 3D-SDN~\cite{yao20183d}. 3D-RelNet~\cite{kulkarni20193d} further extended Factor3D to model object relations. However, these 3D-aware scene representations all rely on annotations of object geometry, which limits their applicability in real scenes with novel objects whose 3D geometry is unseen during training. A few unsupervised models such as Pix2Shape~\cite{rajeswar2020pix2shape}, GQN~\cite{eslami2018neural}, and neural rendering methods~\cite{tewari2020state} have been proposed. However, none of them have yet attempted to model the hierarchy between objects and parts or the relations between multiple scene components. PSGNet builds upon and extends all these ideas to learn a hierarchical, 3D-aware representation without supervision of scene structure.

\section{Methods}

% We first define Physical Scene Graphs mathematically, and then describe the PSGNet architecture.  

\textbf{Physical Scene Graphs.} Informally, a PSG is a vector-labelled hierarchical graph whose nodes are registered to non-overlapping locations in a base spatial tensor. 
More formally, for any positive integer $k$ let $[k] := \{0, \ldots, k-1\}$. 
A \emph{physical scene graph} of depth $L$ is a hierarchical graph $\G = \{(V_l, A_l, E_l, P_l) | l \in [L]\}$ in which $V_l = [|V_l|]$ are layer $l$ vertices, $A_l: V_l \rightarrow \R^{C_l}$ are $C_l$-vector-valued attributes, $E_l$ is the set of (undirected) within-layer edges at layer $l$, and for $l \in [L-1]$, $P_l: V_l \rightarrow V_{l+1}$ is a function defining child-parent edges. 
We also require that for some tensor $S \in \R^{H \otimes W \otimes C}$, $V_0 = [H \cdot W]$ and $A_0[W \cdot i + j] = [i, j] \oplus S[i, j, :]$ for $i \in [H], j \in [W]$, and call $S$ the base of $\G$.
Due to the definition of $\G$, the nodes $V_l$ at any layer define a partition of the base tensor, defined by associating to $v \in V_l$ the set $p(v) = \{(i, j) |  \bigcirc_{l'<l}P_{l'} ((i, j)) = v\}$. 
We call the map $R_l: v \mapsto p(v)$ the \emph{spatial registration} at layer $l$.
An intuitive depiction of an example PSG is shown in Fig. \ref{fig:model}.
A \emph{spatiotemporal} PSG is a union $\bigcup_{t=0}^{t<T} \G^t$ of single time-point PSGs, for which: (i) the first attribute of any node encodes its relative timestep, ie. for $v \in V^t_l$, $A_l(v) = [t] \oplus A^t_l(v)$; and, (ii) for $t_1 < t_2 \in [T]$, we have a (possibly empty) set of \emph{tracking edges} $\T_l^{t_1,t_2} \subset V_l^{t_1} \times V_l^{t_2}$. 

\textbf{PSGNet Architecture.}
PSGNet is a neural network that takes in RGB movies of any length and outputs RGB reconstructions of the same length, along with (i) estimates of  depth and normals maps for each frame, (ii) the (self-supervised) object segment map for each frame, and (iii) the (self-supervised) map of the estimated next-frame RGB deltas at each spatial location in each frame. 
Internally, PSGNet consists of three stages: feature extraction, graph construction, and graph rendering.
In the first stage, a spatially-uniform feature map is created for any input movie by passing it through a Convolutional Recurrent Neural Network (ConvRNN). 
The tensor of features from the lowest (that is, highest spatial resolution) layer of the ConvRNN is then used as the base tensor for constructing a spatiotemporal PSG.  
The PSG is extracted by applying a hierarchical graph constructor that alternates between learnable Graph Pooling and Graph Vectorization. 
Finally, the PSG is the passed through a decoder, which renders graph node attributes (and spatial registrations for the top level of graph nodes) into RGB, depth, normal, segment, and RGB-change maps at each timepoint.
See Fig. \ref{fig:model} for a schematic depiction of the high-level architecture. 
More details of each stage are in the Supplement.

\underline{ConvRNN Feature Extraction.}
A convolutional recurrent neural network (ConvRNN) is a CNN that has been augmented with both local recurrent cells at each layer and long-range feedback connections from higher to lower layers.   
% Such local recurrent and long-range feedback connections are ubiquitous in the primate visual system, where it is hypothesized that they play a role in scene understanding~\cite{}. 
Our feature extraction component is a ConvRNN based on a 5-layer CNN, each layer augmented with an EfficientGatedUnit (EGU), a local recurrent cell that builds in recent insights into local recurrent cell architectures for ConvRNNs~\cite{convrnn}. 
Additionally, long-range feedback connections are introduced from all layers to the first convolutional layer. 
% As ConvRNNs are recurrent networks, they can be unrolled arbitrarily many times.
Unless otherwise specified, we unroll the ConvRNN for $\np=3$ passes.
Given input RGB movie $x$ of shape $(T, H, W, 3)$, our feature extractor takes input $x \oplus \d x$, where $\d x = x[1:] - x[:-1]$ is the one timestep backward differential, creating an input data layer with 6 channels. 
Outputs are taken from the first convolutional layer after each up-down pass, thus forming a tensor $X$ of shape $(T, \np, H, W, C)$, where $H, W, C$ are the output dimensionalities of the Conv 1 layer.
% Further ConvRNN architecture details are given in the supplement.

\underline{Graph Construction.}
Graph construction is a hierarchical sequence of learnable Graph Pooling and Graph Vectorization operations. 
Graph Pooling merges nodes from a previous level to produce the new layer's vertices and child-to-parent edge structure.
Graph Vectorization aggregates and transforms attributes from the nodes at the previous level according to this merging, yielding attribute vectors for the new nodes. 
Together, this produces the next layer of the graph, and the process repeats. 

\textit{Learnable Graph Pooling.}
The idea behind our learnable Graph Pooling operation is that given labeled nodes for a graph layer $l$, the within-layer edges are first computed by thresholding a learnable affinity function on the attribute vectors associated with pairs of nodes. 
Then, the nodes are clustered into groups by propagating putative group identities along these edges, identifying the new nodes for the next graph layer with the resulting clusters. 
The clustering process does not specify the final number of groups (unlike, e.g. $k$-means), allowing the discovery of a variable number of objects/parts as appropriate to different scenes. 

Formally, we define a $k$-dimensional \emph{affinity function} to be any symmetric function $\D_{\phi}: \R^k \times \R^k \rightarrow \R^{\geq 0}$, parameterized by $\phi$, and a \emph{threshold function} to be any symmetric function $\ep: 2^{\R} \times 2^{\R} \rightarrow \R^{\geq 0}$. 
Given a set graph vertices $V_l$, corresponding attributes $A_l: V_l \rightarrow \R^{k_l}$ we then construct the within-layer edges $E_l = \{\{v, w\} | \D_{\phi}(A_l(v), A_l(w)) > \ep(D(v), D(w))\}$, where $D(v) = \{\D_{\phi}(A(v), A(w')) | w' \in V_l\}$.
See below for a description of the specific affinity and threshold functions we use in this work.
For a single timestep PSG, the standard Label Propagation (LP) algorithm~\cite{labelprop} is then run on the graph $G_l = (V_l, E_l)$. LP is an iterative clustering algorithm that initializes by assigning arbitrary unique segment labels to each node in $V_l$. 
At each subsequent iteration, new labels are produced from the previous iteration's labels by assigning to each node the most common label in its radius-1 neighborhood (Fig. \ref{fig:graphbuild}).
The procedure ends after a chosen number of iterations (here ten.)
Let $P_l: V_l \rightarrow [m]$ be defined by the final LP assignment of nodes in $V$ to cluster identities, where $m$ is the number of clusters discovered.
The cluster identities correspond to the new vertices at layer $l+1$ and $m = |V_{l+1}|$. 
For spatiotemporal PSGs, the procedure runs on the set of vertices $V_l = \cup_t V^t_l$, with any edges created across timepoints counted as tracking edges $\T_l$.

\textit{Graph Vectorization.}
The attributes at level $l+1$ are constructed by first computing aggregate labels $A^{\agg}_{l+1}$ from the labels $A_l$ within each new node in $V_{l+1}$, and then transforming the aggregate attributes into new attributes via graph convolution. 
The aggregation is a process of statistical summarization.
For a fixed timepoint and a given vertex $v \in V_{l+1}$, the segment $\Seg[v] := P_l^{-1}(v)$ is the set of nodes in $V_l$ that belong to the cluster associated with vertex $v$.  
The simplest form of summarization would be to take the mean of attributes across nodes in $\Seg[v]$.
To retain more information about each segment, we allow multiple statistical summary functions, aggregated over multiple subregions of $\Seg[v]$.
Specifically, let $\SC[v]$ be a list of subsets of $\Seg[v]$ and $\B$ a list of functions $b: \R^{k_l} \rightarrow \R$.  
Then define $A^{\agg}_{k+1}(v) = [\frac{1}{|s|}\sum_{w \in \text{s}}b(w) \text{ for } s \in \SC[v], b \in \B]$.
We think of $\SC$ as a (discrete) simplicial complex on $\Seg[v]$, and for the purposes of this work take it to include the whole 2D-volume and whole 1D-boundary of $\Seg[v]$, as well as the 4 cardinal volume and 4 cardinal boundary components of $\Seg[v]$ (see Supplement for more details).
We take $\B = \{m: x \mapsto x, v:x \mapsto x^2\}$, corresponding to taking attribute first and second moments respectively. 
See Fig. \ref{fig:model} for graphical illustration of this process.
By construction, since the pixel coordinates $i, j$ were the first two components of $A_0$, the first two attributes of $A^{\agg}_l$ will always be the segment centroid $(c^v_h, c^v_w)$. 
For spatiotemporal PSGs, we ignore the first attribute $t$ in applying the construction, prepending $t$ afterward.
Because the above process relies on integrating across the segments, it produces a fixed-size attribute vector output even when the the size of the segment is not fixed.  
That this type of ``vectorization'' preserves useful geometric information relies on Stokes' theorem for discrete simplicial complexes (see \cite{marsden_dec}).

Additional attributes are created by convolving the aggregated attributes across the $V_{l+1}$ graph vertices. 
In this work, this convolution is implemented by MLPs $H^{\new}_{l+1}$, each with two hidden layers and 100 neurons at each layer. 
(See Supplement for details of implementation.)
The final attribute assignment is given by $A_{l+1}(v) = A^{\agg}_{l+1}(v) \oplus H^{\new}_{l+1}(A^{\agg}_{l+1}(v))$.

%We have now shown how, given data $V_l, A_l, E_l, P_l$ (and $\T_l$) defining the graph at layer $l$, the data $V_{l+1}, A_{l+1}, E_{l+1}$ and $P_{l+1}$ (and $\T_{l+1}$) are computed, thus completing the step in the graph construction hierarchy. 

\underline{Graph Rendering.}
The final step in the process is rendering a PSG back out into feature maps with two types of simple decoder.
% The first takes the spatial registration computed during graph construction and using node attributes as parameters for a kind of ``paints-by-numbers'' texture rendering.
The first takes the spatial registration computed during graph construction and ``paints-by-numbers'' using node attributes.
This procedure generates feature maps that can be compared with ground-truth to generate error signals. 
The second directly constructs new, top-down spatial registrations by using node attributes to ``draw'' shapes in an output image. 
% Outputs of this shape rendering are compared with the previously-computed registration as a self-consistency check. 

\textit{Quadratic Texture Rendering.}
Given the nodes and attributes of layer of a PSG, $V_l, A_l$, together with the layer's spatial registration $R_l$, quadratic texture rendering (QTR) creates a spatial map by inpainting the value of an attribute for node $v$ onto the pixels in $R_l(v)$.   
However, rather than paint uniformly throughout $R_l(v)$, in QTR we paint quadratically.
Let $a, a_{h}, a_{w}, a_{hh}, a_{ww}$, and $a_{hw}$ denote six attribute dimensions from $A_l(v)$.  
Then define the quadratic form $qtr[v](i,j) = a + a_{h} (i - c^v_{h}) + a_{w} (j - c^v_w) + \frac{1}{2} a_{hh} (i - c^v_h)^2 + \frac{1}{2} a_{ww} (j-c^v_w)^2 + \frac{1}{2} a_{hw}(i-c^v_h)(j-c^v_w).$
The associated 2D map is then given by $\QTR^a_l: (i, j) \mapsto qtr[R_l^{-1}(i, j)](i, j)$.  

\textit{Quadratic Shape Rendering.}
We define Quadratic Shape Rendering (QSR) to produce a shape from node attributes, elaborating a procedure developed in ~\cite{convexnets}. 
Let $D$ be a positive integer and $p^d_x, p^d_y, p^d_{\rho}, p^d_{\alpha}, d \in [D]$ be 4$D$ dimensions from $A_l(v)$.  
For each $d \in [D]$, let $qsr^d[v](i, j)$ be the scalar field defined by taking normalized signed distance of point $(i, j)$ to the locus of a 2-D parabola, i.e.
$qsr^d[v](x, y) = \sigma \left ( p^d_{\alpha} [y \cos(p^d_{\rho})-x \sin(p^d_{\rho}) - p^d_x]^2 - [x \cos(p^d_{\rho}) + y \sin(p^d_{\rho}) - p^d_y] \right )$
where $\sigma$ is the standard sigmoid function. 
Let $qsr[v](i, j) = \min_{d \in [D]} qsr^d[v](i, j)).$ 
Define the segment map $\QSR_l: (i, j) \mapsto \arg\max_{v \in V_l} qsr[v](i, j)$ and the segment ``confidence map'' $\QSR^c_l: (i, j) \mapsto \max_{v \in V_l} qsr[v](i, j)$.
In this work, we use $D = 6$.

\textbf{Affinity Functions via Perceptual Grouping Principles.}
The specific choices of edge affinity functions we use encode four core perceptual grouping principles, some of which are inspired by human vision and its development in infants ~\cite{Spelke1990}. 
Described here at a high level, implementation details are found in the Supplement. 

\textit{Principle P1: Feature Similarity.}
Nodes whose features are especially similar should tend to be grouped together. 
% We implement this by first thresholding similarity based on spatial distance, and then computing feature L2 feature affinity.  
We compute feature L2 distance between pairs of features within a fixed spatial window, then threshold each pair by a local average of feature similarity. This creates edges between visual features that are more similar than other pairs in their local neighborhood.

\textit{Principle P2: Statistical Co-occurrence Similarity.}
Nodes whose features co-occur often should tend to be grouped together, even if the features are not similar in the L2 sense. 
To this end, we train a Variational Autoencoder (VAE) to encode feature pairwise differences and use the reconstruction error as an inverse measure of co-occurence frequency and affinity. If a node pair is common, it will be seen more often in training the VAE, so the reconstruction error should be lower (and the affinity higher) than that of a rare node pair.

\textit{Principle P3: Motion-Driven Similarity.}
Nodes that move together should be grouped together, regardless of their visual features. 
This principle is formalized by extending the VAE concept to add tracking edges across nodes at adjacent timepoints, and thus only applies when the input is a movie.

\textit{Principle P4: Learning Similarity from Motion.}
Nodes within a single frame that \textit{look like} they typically move together should tend to be grouped together. 
This principle is implemented by self-supervising an affinity function, parameterized by an MLP, to predict whether the trajectories of two nodes are grouped together by Principle P3.  
This principle generates affinities for static images at test time, but can only be trained from movies. 

\textbf{Loss Functions and Learning.}
Loss functions are applied at two types of locations: (i) in each graph level with learnable affinity functions and (ii) at the end after the rendering, where scene reconstructions are judged for accuracy and segment self-consistency.  

For the reconstruction of RGB and prediction of depth and normal maps, we distinguish at each layer six QTR attribute dimensions for each attribute $\a \in \{R, G, B, z, N_x, N_y, N_z\}$.
The associated loss is $\loss_\a(\th, X) = \sum_{i,j, l, t}(\QTR^{\a(X)}_{\th, l, t}(i, j)-\textbf{gt}_c(X)(i, j))^2$, where $\textbf{gt}_\a(X)$ is the ground-truth value of attribute $\a$ associated with input $X$.
For reconstructing RGB difference magnitude when movies are given, we distinguish a channel $\Delta$ (and its associated QTR attributes), self-supervising on $|\sum_{r,g,b} (\d X)^2|$ for ground-truth.
We also compute losses for QSR predictions, taking the softmax cross-entropy between the shape map for each predicted object and the self-supervising ``ground truth'' indicator function of whether a pixel is in that object's segment, i.e. $\loss_{\QSR}(\th, X) = \sum_{t, i,j, v \in V_L(X_t)} \SMCE(qsr[v](i, j), \id[(i, j) \in R_L(v)])$.

To compute a projective 2D-3D self-consistency, we treat our system as a pinhole camera.
We distinguish two additional attributes $x$ and $y$ at each graph layer and compute the loss $\loss_{\proj}(X) = \sum_{v \in V_l(X), t}||\proj[a^t_x(v), a^t_y(v), a^t_z(v)] - (c^{v, t}_h, c^{v,t}_w)||_2$ where $a^t_z(v)$ is the predicted depth (from the depth-supervised channel $z$) and $\proj: \R^3 \rightarrow \R^2$ denotes the pinhole camera perspective projection. The camera focal length parameter is provided on a per-dataset basis in training.

\section{Experiments and Analysis}

\textbf{Datasets, Baselines, and Evaluation Metrics.}
We compare PSGNet to recent CNN-based object discovery methods based on the quality of the self-supervised scene segmentations that they learn on three datasets.  
\Prim\ is a synthetic dataset of primitive shapes (e.g. spheres, cones, and cubes) rendered in a simple 3D room. 
\Play\ is a synthetic dataset of objects with complex shapes and realistic textures, such as might be found in a child's playroom (e.g. animals, furniture, and tools), rendered as movies with object motion and collisions.
Images in \Prim\ and \Play\ are generated by ThreeDWorld (TDW), a general-purpose, multi-modal virtual world simulation platform built on Unity Engine 2019. TDW is designed for real-time near-photorealistic rendering of interior and exterior environments, with advanced physics behavior that supports complex physical interactions between objects.
% \Play motion dynamics are generated by the NVIDIA FleX physics engine. 
\Gib\ is a subset of the data from the Gibson1.0 environment~\cite{stanford2d3d}, composed of RBG-D scans of inside buildings on the Stanford campus. 
All three datasets provide RGB, depth, and surface normal maps used for model supervision, as well as segmentation masks used to evaluate predicted unsupervised segmentations. 
Details of the datasets and evaluation metrics are provided in the Supplement.

We compare PSGNet to three neural network baselines --- MONet \citep{monet}, IODINE \citep{iodine}, and OP3 \citep{op3} --- and one non-learned baseline, Quickshift++ (Q++) \cite{quickshiftpp}, which is given ground truth RGB, depth, and normals maps as input.
All three neural network baselines have similar CNN-based architectures and hyperparameters and are trained with Adam \citep{kingma2015adam} as described in the Supplement.

\begin{table}[t]
    \centering\small
    \setlength{\tabcolsep}{4.5pt}
    \caption{Performance of models on TDW-Primitives, TDW-Playroom, and Gibson test sets after training on each set. OP3 and PSGNetM were not trained on Gibson or Primitives as these have static test sets. Quickshift++ (Q++) receives ground truth depth and normals maps as input channels in addition to RGB.}
    \begin{tabular}{lcccccccccc}
    \toprule
    & \multicolumn{3}{c}{Primitives} & \multicolumn{3}{c}{Playroom} & \multicolumn{4}{c}{Gibson}\\
    \cmidrule(lr){2-4}\cmidrule(lr){5-7}\cmidrule(lr){8-11}
    Models & Recall & mIoU & BoundF & Recall & mIoU & BoundF & Recall & mIoU & BoundF & ARI \\
    \midrule
    MONet & 0.35 & 0.40 & 0.50 & 0.28 & 0.34 & 0.46 & 0.06 & 0.12 & 0.15 & 0.27 \\
    IODINE & 0.63 & 0.54 & 0.57 & 0.09 & 0.15 & 0.17 & 0.11 & 0.15 & 0.14 & 0.30 \\
    Q++ (RGBDN) & 0.55 & 0.54 & 0.62 & 0.50 & 0.53 & 0.65 & 0.20 & 0.20 & 0.24 & 0.45 \\
    OP3 & - & - & - & 0.24 & 0.28 & 0.31 & - & - & - & - \\
    PSGNetS & \textbf{0.75} & \textbf{0.65} & \textbf{0.70} & 0.64 & 0.57 & 0.66 & \textbf{0.34} & \textbf{0.38} & \textbf{0.37} & \textbf{0.53} \\
    PSGNetM & - & - & - & \textbf{0.70} & \textbf{0.62} & \textbf{0.70} & - & - & - & - \\
    \bottomrule
    \end{tabular}
    \vspace{-5pt}
    \label{tab:performance_tdw}
\end{table}

% \begin{table}
%     \centering
%     \caption{Performance of models Gibson test set.}
%     \begin{tabular}{lllll}
%     \hline\noalign{\smallskip}
%     Model & Recall & mIoU & BoundF & ARI \\
%     \noalign{\smallskip}
%     \hline
%     \noalign{\smallskip}
%     MONet & 0.06 & 0.12 & 0.15 & 0.27 \\
%     IODINE & 0.11 & 0.15 & 0.14 & 0.30 \\
%     Q++ (RGBDN) & - & - & - & 0.33 \\
%     PSGNetS & 0.34 & 0.38 & 0.37 & \textbf{0.53} \\
%     \end{tabular}
%     \label{tab:performance_gibson}
% \end{table}

\begin{figure}[t]
\centering
    \includegraphics[width=1.0\textwidth]{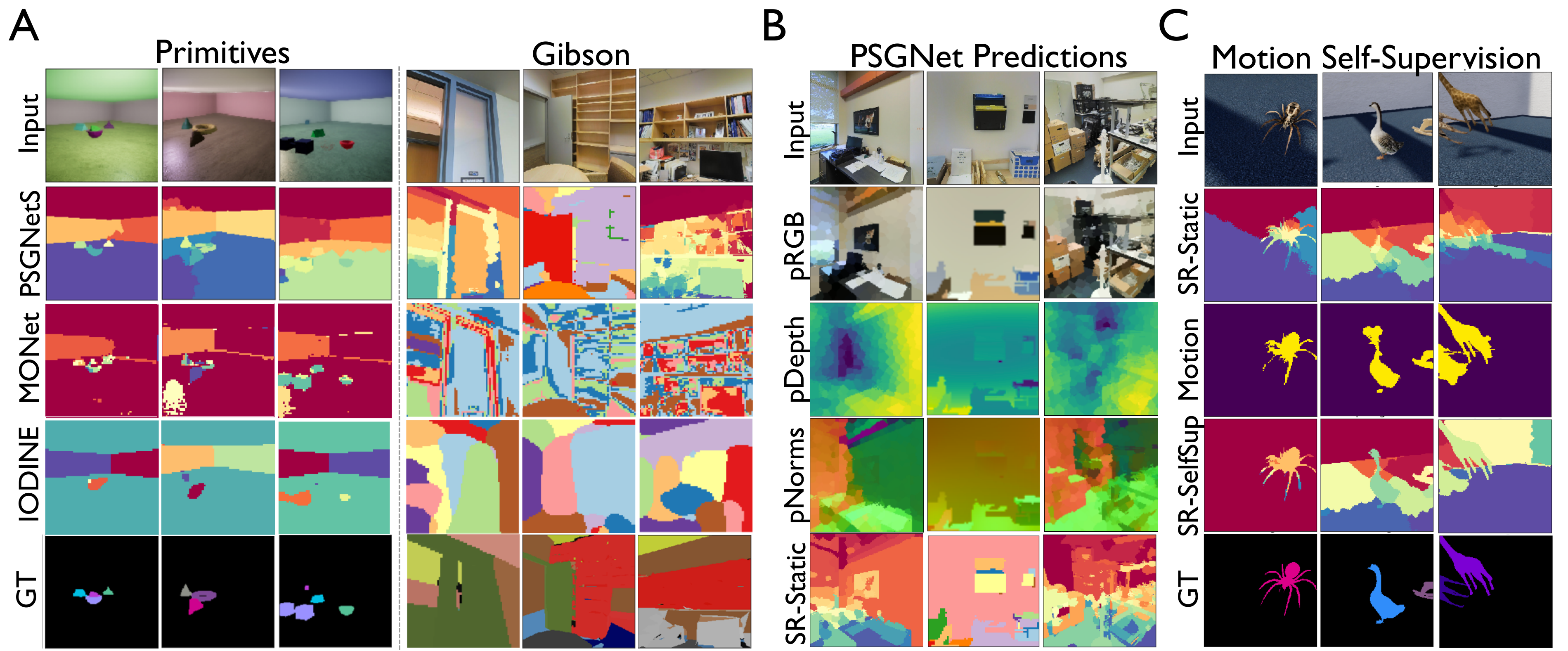}
\vspace{-15pt}
\caption{Visualizing the scene decompositions of PSGNets. (\textbf{A}) The predicted scene segmentations of baselines and PSGNetS. The PSGNetS segmentation is the top-level spatial registration (SR) from the PSG. (\textbf{B}) Rendering of the lowest level PSGNetS node attributes and the top-level SR on Gibson images. (\textbf{C}) Comparing the static SR (SR-Static) and the SR learned by motion-based self-supervision (SR-SelfSup) in PSGNetM. These SRs are predicted from single frames, not movies, at test time; row two (Motion) indicates where motion \textit{would} be detected during training, according to a separate evaluation of these images as frames of movies.}
\label{fig:qualitative}
\vspace{-20pt}
\end{figure}

\textbf{Static Training.}
We first compared models using static images from \Prim\ and \Gib, with RGB reconstruction and depth and surface normal estimation as training signals. 
All models were given single RGB frames as input during training and testing. 
On \Prim, PSGNetS (PSGNet with static perceptual grouping only, see Supplement) substantially outperformed alternatives, detecting more foreground objects and more accurately predicting ground truth segmentation masks and boundaries than MONet \citep{monet} or IODINE \citep{iodine} (Table \ref{tab:performance_tdw}). 
IODINE and PSGNetS also outperformed the un-learned segmentation algorithm Q++ \citep{quickshiftpp}.
Only PSGNetS learned to plausibly decompose real static scenes from the \Gib\ dataset.
The per-segment mask and boundary detection, as well as Adjusted Rand Index (ARI) -- which measures the rate of predicted pixel \textit{pairs} correctly falling into the same or distinct ground truth segments -- were nearly two-fold higher for PSGNetS than either learned baseline.
 
% Gibson reveals limitations of static segmentation
While PSGNetS makes characteristic grouping errors on Gibson images, its scene decompositions are broadly plausible (Fig. \ref{fig:qualitative}A, B.) 
In fact, many of its ``errors'' with respect to the segment labels illustrate the difference between a physical scene representation and a semantic description:
what is labeled a single bookcase in the dataset, for instance, actually holds many visually and physically distinct objects that PSGNetS identifies (Fig. \ref{fig:qualitative}A, rightmost column.)
Reconstruction of appearance and surface geometry (Fig. \ref{fig:qualitative}B) cannot resolve these ambiguities because objects can have arbitrarily complex shapes and textures.
Instead, true ``objectness'' must be at least partly learned from a dynamical principle: whether a particular element can move independently of the rest of the scene. 

\textbf{Motion-Based Training.}
We therefore trained PSGNetM, a PSGNet equipped with the motion-based grouping principles P3 and P4, on four-frame movies from the \Play\ dataset (see Supplement.)
Tested on \textit{static} images, PSGNetM produced more accurate scene segmentations than those of PSGNetS by learning to group visually distinct parts of single objects (Fig. \ref{fig:qualitative}C.)
PSGNetS in turn achieved higher performance than either MONet or IODINE when trained and tested on \Play.
OP3, a recent algorithm that encodes and decodes movie frames with the IODINE architecture, but also learns a simple graph neural network dynamics model on the inferred latent factors to reconstruct \textit{future} images, could in principle learn to segment static scenes better from motion observed during training \citep{op3}.
While OP3 outperforms the static IODINE architecture, it is not nearly enough to make up the gap between the CNNs and the PSGNet models. 

\textbf{Learning Efficiency and Transfer.}
Together, these results suggest that PSGNet has better inductive biases than CNN-based methods for unsupervised detection of complex objects and decomposition of real scenes.
Two further observations support this claim.
First, PSGNet models are more than two orders of magnitude more sample efficient in learning to decompose scenes.
When training on either \Prim\ or \Play, segmentation performance on held-out validation images saturated within two training epochs for PSGnetS and PSGNetM but only after 200-400 epochs for the CNNs on \Prim\ or 20-100 epochs on \Play\ (though CNNs badly underfit on the latter; Table \ref{tab:learning}.)
Second, PSGNet models trained on one synthetic dataset largely transferred to the other -- despite zero overlap between the object models in each -- while the baselines did not.
Test set recall on \Prim\ of \Play-trained PSGNetM was 0.56 (80\% of its within-distribution test performance), whereas that for \Play-trained OP3 was 0.10 (42\% transfer); transfers from \Prim\ to \Play\ were 51\% and 29\% for PSGNetS and MONet, respectively (see Supplement.)
These findings suggest that disentangled scene properties and pairwise feature relationships -- encoded in PSGNet graph nodes and affinity functions, respectively -- are more efficient to learn, and more generally applicable across scenes, than the per-object segmentation masks predicted by the CNN models.
In the Supplement we discuss typical failure modes for each model.

\begin{table}[t]
    \setlength{\tabcolsep}{2pt}
    \centering\small
    \caption{Performance of Ablated PSGNetM variants on TDW-Playroom.}
    %\vspace{-5pt}
    \begin{tabular}{lccccccccccccc}
    \toprule
    Ablation & Base & NoLoc & NoFB & FF & UNetL & NoBo & NoVar & NoGC & NoQR & NoDN & NoAll & NoDN* & BAff \\
    \midrule
    Recall & \textbf{0.70} & 0.59 & 0.52 & 0.47 & 0.55 & 0.63 & 0.61 & 0.57 & 0.59 & 0.41 & 0.47 & 0.65 & 0.05 \\
    mIoU & \textbf{0.62} & 0.54 & 0.50 & 0.48 & 0.52 & 0.57 & 0.56 & 0.53 & 0.54 & 0.44 & 0.47 & 0.58 & 0.26 \\
    BoundF & \textbf{0.70} & 0.55 & 0.50 & 0.49 & 0.53 & 0.59 & 0.58 & 0.54 & 0.55 & 0.47 & 0.46 & 0.59 & 0.25 \\ 
    \bottomrule
    \end{tabular}
    \vspace{-10pt}
    \label{tab:ablations}
\end{table}

\begin{figure}[t]
\centering
    \includegraphics[width=1.0\textwidth]{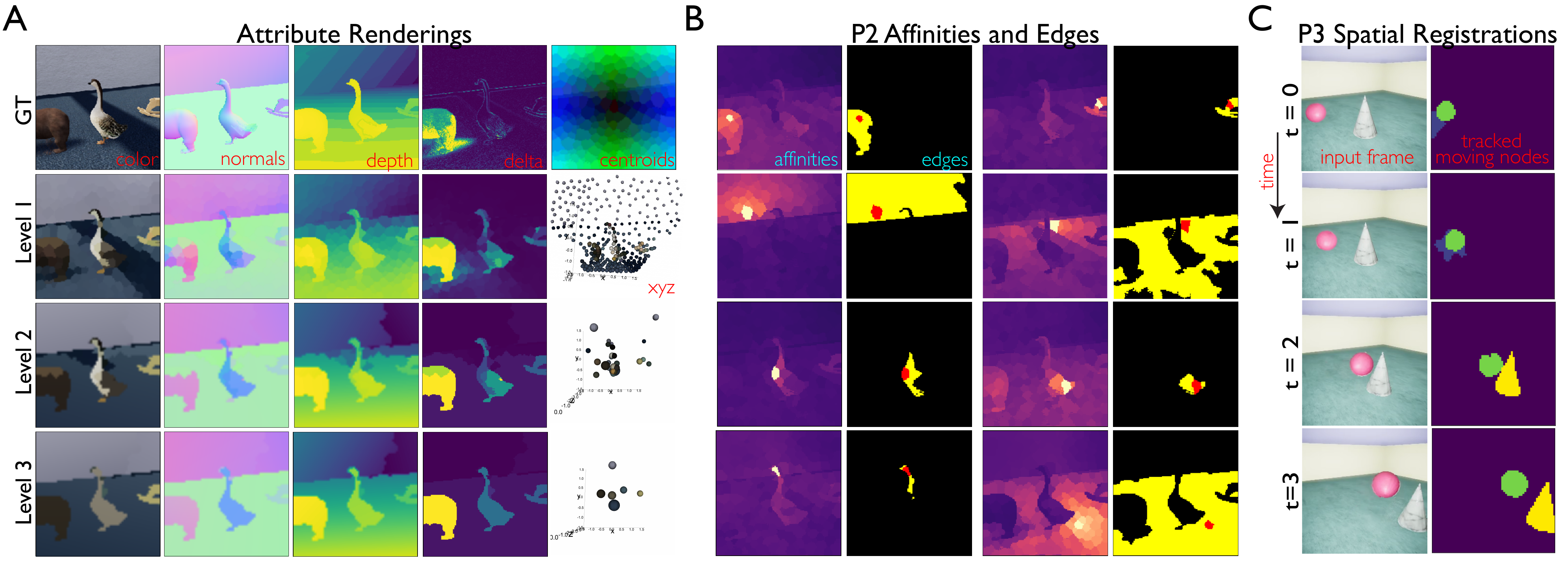}
\vspace{-17pt}
\caption{Visualizing the predicted components of a PSG. (\textbf{A}) The QTRs of different PSG levels for the input image in the top left. \textbf{Delta} is the magnitude of RGB change relative to the prior frame. Nodes plotted by their (x,y,z) attributes are colored by their RGB attributes, with nodes that cover <25 pixels hidden for clarity. (\textbf{B}) Eight examples of learned P2 affinities (left) and binary edges (right) between the nodes marked in yellow (left) or red (right) and other nodes. Warmer colors indicate higher affinity. (\textbf{C}) Graph nodes and spatial registrations inferred from motion (P3 affinity.) Segments of the same color across different frames belong to the same spatiotemporal node.}
\label{fig:psgviz}
\vspace{-20pt}
\end{figure}

\textbf{Ablations of the PSGNetM architecture.}
Deleting local recurrence (NoLoc), long-range feedback (NoFB), or both (creating a feedforward network, FF) from the PSGNetM ConvRNN increasingly hurt segmentation performance on \Play\ (Table \ref{tab:ablations}.)
Moreover, a "U-Net-like" ConvRNN (UNetL) in which feedforward and upsampled recurrent features are simply summed instead of combined over time also performed worse than the base model. 
This is likely because the Conv1 features that initialize graph construction are over-smoothed by recurrent convolutions.
Models without aggregation of segment border attributes (NoBo), aggregation of feature variances (NoVar), graph convolutions between node pairs (NoGC), or quadratic (rather than constant) texture rendering (NoQR) also performed substantially worse than the base model.

Removing depth and normals decoding (NoDN) without otherwise changing the PSGNetM architecture badly hurt performance; yet this model still outperformed the CNN baselines \textit{with} depth and normals supervision (cf. Table \ref{tab:performance_tdw}), indicating that PSGNets' better segmentation does not stem only from learning explicit scene geometry.
The performance of NoDN was comparable to that of a model with depth and normals decoding restored, but all ConvRNN recurrence and optional Graph Construction components ablated (NoAll).
This level of performance may therefore represent the benefit of perceptual grouping-based over CNN-based inductive biases, while the additional ConvRNN and Graph Construction components mostly allow PSGNetM to learn how to integrate geometric cues into scene decomposition.
Without depth and normals supervision, PSGNets perform a much easier \textit{RGB-autoencoding} task, so that additional architectural complexity and parameters may lead to overfitting.
Consistent with this interpretation, a new hyperparameter search over ``RGB-only'' PSGNet architectures identified a relatively high-performing \textit{feedforward} model (NoDN*) with no ConvRNN or graph convolutional recurrence.
This model nevertheless falls well short of the base PSGNetM trained with geometric cues.
Together, these results suggest that PSGNets benefit from but do not require geometric supervision.
On the other hand, simpler architectures (e.g. NoAll) appear to \textit{underfit} the combination of appearance and geometric cues, such that their segmentation performance is much worse than a hyperparameter-optimized autoencoding architecture (NoDN*).

Last, we converted the P2 affinity function from a VAE to an MLP that predicts only whether two nodes are adjacent in the input image.
Using this binary (rather than vector-valued) affinity signal greatly reduced performance (BAff, Table \ref{tab:ablations}), indicating that segment adjacency alone is not a reliable signal of feature co-occurrence in \Play;
however, both affinity signals may be stronger when training on more diverse datasets \citep{isola2015learning}.

\textbf{Visualizing learned PSGs.}
Renderings from the nodes of an example PSG closely resemble the visual features they were trained to predict (Fig. \ref{fig:psgviz}A.) 
Because of quadratic rendering, a single higher-level node can represent a large, smoothly changing scene element, such as a floor sloping in depth.
This often reflects underlying physical structure better than other 3D scene representations, like dense meshes \citep{gkioxari2019mesh}: 
a flat wall need not be encoded as jagged, and a sphere's roundness need not depend on how many surface points are supervised on.
PSG edges do not have observable supervision signals to compare to.
However, rendering the affinity strengths and edges from a chosen node to other nodes (Fig. \ref{fig:psgviz}B) shows that the strongest static (P2) relationships are within regions of similar visual or geometric attributes, as expected from their frequent co-occurrence.
Regions that differ only in some attributes (such as lighted and shadowed parts of the floor) can still have high affinity -- even spreading "around" occluding objects -- as this common arrangement can be reconstructed well by the P2 VAE.
When shown a movie, P3 affinities group nodes that are moving together.
This determines the tracking of moving nodes, such that nodes in different frames are identified with a single spatiotemporally extended object (Fig. \ref{fig:psgviz}C.)
Using these moving segments to self-supervise the P4 affinity function allows for grouping visually distinct scene elements, but also can overgroup similar-looking, nearby objects (Fig. \ref{fig:qualitative}C.)
The Supplement contains more PSG examples.

\begin{figure}[t]
\begin{center}
    \includegraphics[width=1.0\textwidth]{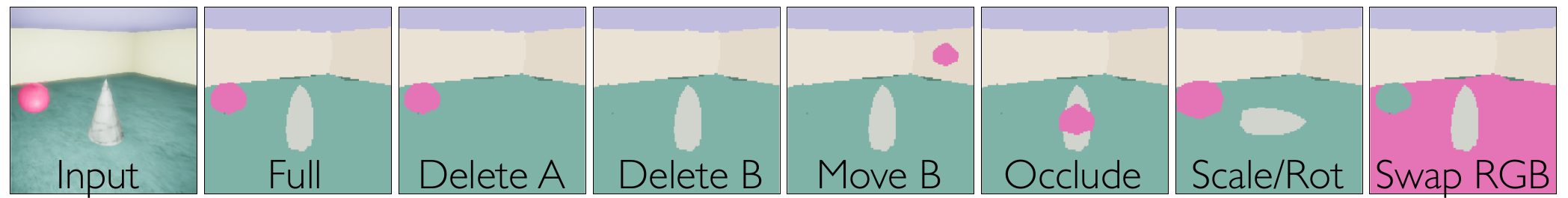}
\end{center}
\vspace{-12pt}
\caption{Symbolic manipulation of a PSG. An example PSG is manually edited before shape rendering (here each shape is colored by its node's RGB attributes.) Deleting chosen nodes (DeleteA/B), moving them to new 3D locations (MoveB and Occlude), altering shape attributes (Scale/Rot), or swapping two nodes' colors (Swap RGB) all change the graph rendering as predicted relative to the original (Full). These graph edits are implemented through symbolic instruction, i.e. code that picks out one or more nodes and assigns new values to components of their attribute vectors.}
\label{fig:compositional}
% \vspace{-20pt}
\end{figure}

\textbf{The symbolic structure of PSGs.} 
We illustrate the symbolic properties of a PSG by "editing" the nodes of its top level and observing effects on its rendering (Fig. \ref{fig:compositional}.) 
If a PSG correctly represents a scene as discrete objects and their properties, then manual changes to the nodes should have exactly predictable effects.
This is in fact what we observe.
Deleting a node from the graph removes only that node's rendered shape, and changing an attribute (e.g. 3D position, shape, or color) affects only that property.
The "filling in" of vacated regions indicates that a node's representation extends through visually occluded space.
Because PSGs support symbolic instruction -- here coded manually -- future work will allow agents to \textit{learn} how to physically manipulate scene elements to achieve their goals.
Furthermore, an agent equipped with a PSGNet-like perceptual system, observing the dynamics of its internal PSG representation, could learn a form of \textit{intuitive physics}: 
a model of which changes to the PSG are physically possible or likely. 
Given the promising results of applying graph neural networks to supervised dynamics prediction problems \cite{mrowca2018flexible, li2018learningParticle, li2020visual, sanchez2020learning}, the graph-structured latent state of PSGNet seems well-poised to connect visual perception with physical understanding.

% \input{MainText/data_and_eval}
% \input{MainText/baselines_and_controls}
% \input{MainText/results}
% \section{Old Experiments}
% See OldText.
% \input{OldText/results_old}

\section{Acknowledgements}
This work was funded in part by the IBM-Watson AI Lab.
D.M.B. is supported by an Interdisciplinary Postdoctoral Fellowship from the Wu Tsai Neurosciences Institute and is a Biogen Fellow of the Life Sciences Research Foundation.
D.L.K.Y is supported by the McDonnell Foundation (Understanding Human Cognition Award Grant No. 220020469), the Simons Foundation (Collaboration on the Global Brain Grant No. 543061), the Sloan Foundation (Fellowship FG-2018- 10963), the National Science Foundation (RI 1703161 and CAREER Award 1844724), the DARPA Machine Common Sense program, and hardware donation from the NVIDIA Corporation.
We thank Google (TPUv2 team) and the NVIDIA corporation for generous donation of hardware resources.

{\small
% \bibliographystyle{plainnat}
% \bibliography{reference,psgnet}

}
\newpage
\setcounter{section}{0}
\setcounter{figure}{0}
\setcounter{table}{0}
\renewcommand{\thesection}{S\arabic{section}}  
\renewcommand{\thefigure}{S\arabic{figure}}  
\renewcommand{\thetable}{S\arabic{table}}  

\section{PSGNet Architecture}
The PSGNet architecture consists of three stages: feature extraction, graph construction, and graph rendering.
In the first stage, a spatially-uniform feature map is created for any input movie by passing it through a Convolutional Recurrent Neural Network (ConvRNN) $\F_{\Th_0}$. 
The tensor of feature activations from one convolutional layer of the ConvRNN is then used as the base tensor for constructing a spatiotemporal PSG.  
Taking these features as the ``Level-0'' nodes, the higher levels of a PSG are built one at a time by applying a learned Graph Constructor $\GC_{\Th_1}$. This stage itself contains two types of of module, \textit{Graph Pooling} and \textit{Graph Vectorization.}
Thus the final PSG representation of an input movie has $L+1$ levels after applying $L$ (Pooling, Vectorization) pairs. 
Finally, the PSG is the passed through a decoder $\Rndr$, which renders graph node attributes (and spatial registrations for the top level of graph nodes) into RGB, depth, normal, segment, and RGB-change maps for each frame of the input movie.

Formally, then, we define the parameterized class of neural networks $\PSGN_{\Th}$ as functions of the form
\begin{equation}
\PSGN_{\Theta} = \Rndr \circ \GC_{\Theta_1} \circ \F_{\Theta_0},
\end{equation}
where $\F$ is the ConvRNN-based feature extractor, $\GC$ is the graph constructor, $\Rndr$ is the graph rendering, and $\Th = \Th_0 \cup \Th_1$ are the learnable parameters of the system. 

Note that the decoder does not have any learnable parameters of its own: it takes as input only the spatial registrations produced by Graph Pooling and the node attributes produced by Graph Vectorization, using them to ``paint-by-numbers'' (Quadratic Texture Rendering, $\QTR$) or ``draw shapes'' (Quadratic Shape Rendering, $\QSR$) as described in the main text.
This strong constraint on decoding is what allows PSGNet to learn \textit{explicit} representations of scene properties.
Without additional parameters to convert a latent code into a rendered image, the latent codes of a PSG (i.e. the node attribute vectors) are optimized to have the same encodings as their (self-)supervision signals -- color, depth, surface normal vectors, etc.

Below we describe each component of the PSGNet architecture in detail and give further background for its motivation. We then give the specific hyperparameter choices used in this work.

\subsection{ConvRNN Feature Extraction}

A convolutional recurrent neural network (ConvRNN) is a CNN augmented with both local recurrence at each layer and long-range feedback connections from higher to lower layers.   
Such local recurrent and long-range feedback connections are ubiquitous in the primate visual system, where it is hypothesized that they play a role in object recognition and scene understanding \cite{kar2019evidence, tang2018recurrent, van2020going}. 
Large-scale neural network implementations of ConvRNNs optimized for ImageNet categorization have been shown achieve performance gains on a per-parameter and per-unit basis, and make predictions of neural response dynamics in intermediate and higher visual cortical areas \cite{convrnn}. 
Moreover, models with both local and long-range recurrence are substantially more parameter-efficient than feedforward CNNs on a variety of perceptual grouping tasks, which suggested that multiple forms of recurrence could provide useful inductive biases for scene decomposition \cite{kim2019disentangling, linsley2018learning,  linsley2020stable, george2017generative}.

ConvRNNs prove useful in our application because they naturally integrate high- and low-level visual information, both of which are critical to understand scene structure.   
The bottom-up-and-top-down ConvRNN architecture is related to the standard U-Net structure (e.g. \cite{monet}.)
Long-range feedback in a ConvRNN plays a role analogous to the upsampling layers of a U-Net, while locally recurrent cells are similar to ``skip-connections'' in that they combine features of the same spatial resolution.
However, the ConvRNN provides more flexibility than a U-Net in terms of how information from multiple layers is integrated.
First, it can be unrolled for any number of ``passes'' through the full network (the operation of each feedforward, locally recurrent, and feedback connection), with more passes producing more nonlinear feature activations at each layer.
Second, we can take advantage of this change in features over time by performing different stages of PSG construction after differing numbers of passes.
For example, the initial construction of Level-1 nodes from ConvRNN features (Level-0 nodes) yields the sharpest partition of the scene into ``superpixel''-like segments when done after a single ConvRNN pass;
this is because further perturbations to the ConvRNN features by recurrent local and long-range convolution tend to smooth them out, degrading the precise boundaries between scene elements.
On the other hand, features obtained after multiple passes have integrated more mid- and high-level information about the scene -- including across the larger spatial receptive fields found in higher ConvRNN layers -- making them more accurate at predicting mid- and high-level PSG attributes, like depth and surface normals (data not shown.)

\textbf{Implementation.}
Formally, let $F^p_k$ be the ConvRNN feature activations at layer $k$ of the backbone CNN after pass number $p$.
% long-range fb
Long-range feedback connections combine features from a higher layer $k+k'$ with this lower, potentially spatially larger layer $k$. This takes the form
\begin{equation}
    \tilde{F}^p_k = ReLU(U_k^{k'} \ast Resize(F^{p-1}_{k+k'})),
\end{equation}
where $U_k^{k'}$ is a convolution kernel and $Resize$ is bilinear upsampling to the resolution of the target layer.
On the initial pass ($p=0$), ${F}^{p-1}_{k+k'}$ are defined to be a tensor of zeros.
% The feedback features $\tilde{F}^p_k$ are then combined with feedforward and local recurrent features at that layer.

Local recurrent connections combine features within a CNN layer, which necessarily have the same spatial dimension:
\begin{equation}
    F^p_k = Combine_{\{W_k, U_k\}}(F^p_{k-1},\ F^{p-1}_k, \ \tilde{F}^p_k),
\end{equation}
where $W_k$ and $U_k$ are, respectively, the feedforward and recurrent convolution kernels at that layer, $\tilde{F}^t_k$ are any features produced by feedback to that layer, and $Combine(a, b, c)$ is a nonlinear function that combines the convolved features, such as $ReLU(a + b + c)$.
The functional form of $Combine$ defines a ``Local Recurrence Cell'' (such as a modified Vanilla RNN or LSTM Cell \cite{convrnn}.)
As with feedback, locally recurrent features $F^{p-1}$ are defined to be zero on the initial pass.

% summarize
Thus, a ConvRNN architecture is fully specified by its backbone CNN, its list of feedback connections $[(k_0,k_0 + k'_0), (k_1,k_1 + k'_1), \dots]$, the structure of its $Combine$ functions, and hyperparameters of all convolution kernels. 

\textbf{In this work.}
Here we use a backbone CNN with $5$ feedforward layers. 
The layers have $(40, 64, 96, 128, 192)$ output channels, respectively, each followed by $2\times2$ max pooling with stride $2$ to downsample the spatial dimensions by a factor of two per layer.
There are feedback connections from all higher layers to the first layer, whose feature activations become the Level-0 PSG nodes.
The locally recurrent ``Cell'' installed at each layer is what we call an EfficientGatedUnit (EGU), as it was inspired by the parameter-efficient, \textit{feedforward} architecture EfficientNet \cite{tan2019efficientnet} and designed to satisfy two properties: (i) learned gating of convolutional inputs, and (ii) pass-through stability, i.e. on the initial pass through the ConvRNN, outputs from the convolution operation are unaffected by the recurrent cell before being passed to the next layer. 
Together, these properties mean that local recurrence produces a dynamical perturbation to the feedforward activations, which was previously found to help performance on ImageNet classification \cite{convrnn}. 
The EGU $Combine$ function has the form 
\begin{equation}
    \overline{F}^p_k = F^{p-1}_k + \tilde{F}^p_k + ReLU(U_k \ast (W^{in}_k \ast F^p_{k-1} + F^{p-1}_k)),
\end{equation}
\begin{equation}
    F^p_k = W^{out}_k \ast [\sigma(W^e_k \ast ReLU(W^r_k \ast <\overline{F}^p_k>)) \odot \overline{F}^p_k],
\end{equation}
where equation $(5)$ is a ``Squeeze-and-Excitation'' gating layer \cite{hu2018squeeze}:
it takes mean of input features across the spatial dimensions, $<\overline{F}>$, reduces and expands their channel dimensions with the $1\times1$ kernels $W^r_k$ and $W^e_k$, then multiplicatively gates the input $\overline{F}$ with the sigmoid function $\sigma$ of this spatially broadcast tensor.
The final layer output is a feedforward convolution of this tensor with the kernel $W^out_k$.
Other than $W^r_k$ and $W^e_k$, all feedforward kernels $W_k$ are $3\times3$; All local recurrent kernels $U_k$ are $5\times5$; and all feedback kernels $U_k^{k'}$ are $1\times1$.
The ``efficiency'' of this cell comes from making the kernels $W^{in}_k$ and $U_k$ \textit{depth-separable}, i.e. a convolution that operates per feature channel with the full spatial kernel dimensions followed by a $1\times1$ convolution that mixes channels.
Following the original EfficientNet structure, we have the $W^{in}$ increase the channel dimension by a factor of $6$ at each layer and the $W^r$ reduce it by a factor of $4$. 
This fully determines the required output channel dimensions of all other convolution kernels.

% Total input/output
Given a RGB movie $x$ written in channel-minor form with shape $(T, H_0, W_0, 3)$, we form the backward temporal difference $\delta x = x[1:T] - x[0:T-1]$ (where a block of 0s is padded on the end of $x[1:T]$ and the beginning of $x[0:T-1]$), and then form the channel-wise concatenation $\Delta(x) = x \oplus \delta x$, so that the input data layer has 3*2 = 6 channels. 
The ConvRNN is then run independently on each input frame of $\Delta(x)$ for $n_{\text{unroll}}=3$ passes.
We take outputs from the first convolutional layer in $\F_{\Th_0}(\Delta(x))$ after each up-down pass, forming a tensor of shape $(T, n_{\text{unroll}}, H, W, C)$, where $H, W, C$ are the output dimensionalities of the Conv1 layer.
This is used as input to the graph constructor.
PSGNetS is trained on static images (i.e. movies of length $T=1$) while PSGNetM is trained on movies of length $T=4$;
either model can be evaluated on movies of arbitrary length.
All ConvRNN parameters are optimized end-to-end with respect to the $\QSR$ and $\QTR$ losses on the final PSG.

\subsection{Learned Perceptual Grouping and Graph Pooling}

\begin{figure}[t]
\centering
    \includegraphics[width=1.0\textwidth]{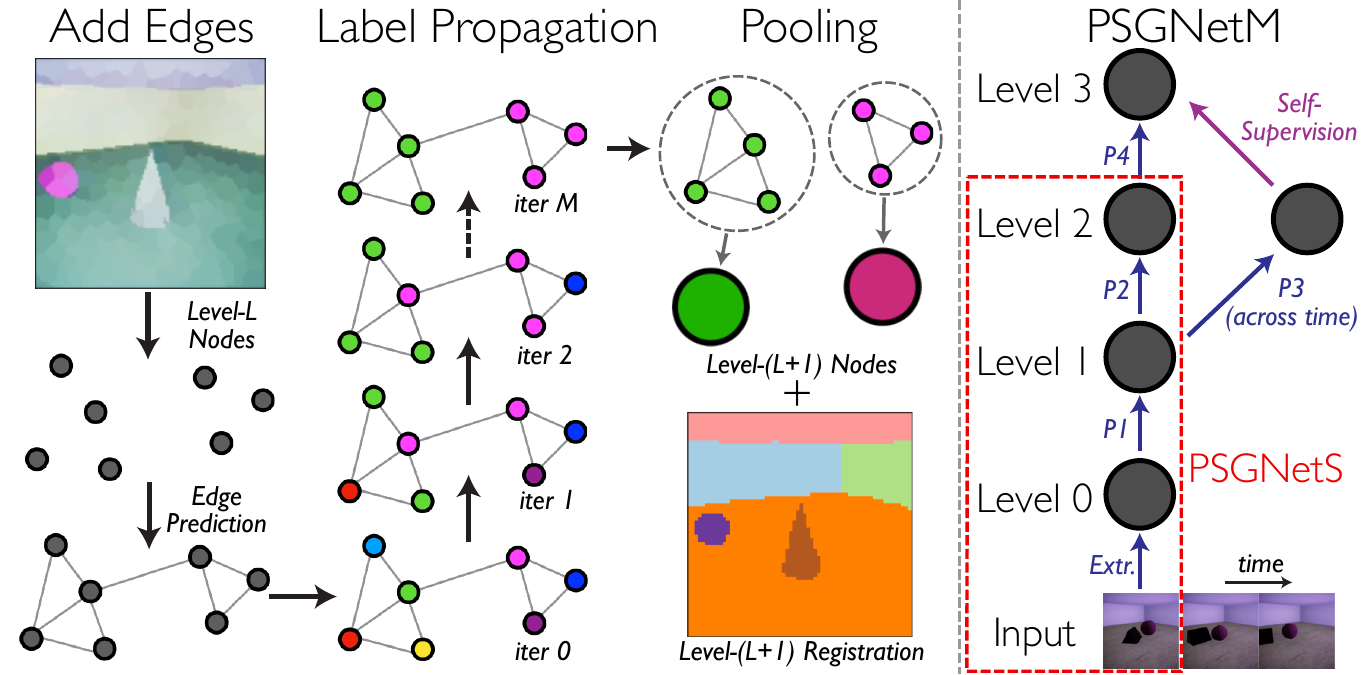}
\vspace{-15pt}
\caption{Schematics of the generic process for building a new PSG level and of the actual PSGNet architectures used in this work. (\textbf{Left}) To build nodes $V_{l+1}$, existing nodes $V_l$ (here shown as their spatially registered color rendering) are passed through pairwise affinity and thresholding functions to \textit{Add Edges} $E_l$. The resulting graph $(V_l, E_l)$ is clustered according to \textit{Label Propagation}; here nodes are colored by their assigned label (cluster index) at each iteration, which is updated by taking the most common label among graph neighbors. (Ties are resolved randomly.) After $M$ iterations, the kernels for \textit{Pooling} nodes are defined by the final cluster assignments. This determines the child-to-parent edges $P_l$ and the new spatial registration $R_{l+1}$. Graph Vectorization, which constructs the new attributes $A_{l+1}$, is not shown here. (\textbf{Right}) The set of operations and perceptual grouping principles used to build PSGs in PSGNetM and PSGNetS (red outline.) Other than grouping by motion-in-concert (P3), all grouping constructs per-frame PSG nodes $V^t_l$. P3 builds Level 2M nodes that extend temporally across all input frames. \textit{Extr} indicates ConvRNN feature extraction.}
\label{fig:graphbuild}
\vspace{-15pt}
\end{figure}

The features extracted from ConvRNN layer $k$ after a chosen number of passes $p$ are considered, for each input movie frame $t$, the Level-0 vertex/node sets $V^t_0$.
Thus $|V^t_0| = H \cdot W$ and the attribute map $A_0: V_0 \rightarrow \R^{C}$ is defined as 
\begin{equation}
    A_0(v) = A_0(W\cdot i + j) = [t] \oplus [i,j] \oplus F_k[t,p,i,j,:],
\end{equation}
that is, simply indexing into each spatial position of the base feature activation tensor and prepending the movie time step and spatial coordinates.
At this stage, there are no within-layer edges or parent edges (because there are no higher-level nodes) and the spatial registration $R_0$ is the trivial partition of singleton nodes.
As described in the main text, building a new set of child-to-parent edges $P_0$ from $V_0$ -- or more generally, $P_l$ from $V_l$ -- requires three computations: 
(1) a function $D_{\phi}$ (which may have learnable parameters) that assigns a nonnegative affinity strength to each pair of nodes $(v,w)$;
(2) a thresholding function $\ep$ (which has no learnable parameters) that converts real-valued affinities to binary within-layer edges, $E_l$;
and (3) an algorithm for clustering the graph $(V_l, E_l)$ into a variable number of clusters $|V_{l+1}|$, which immediately defines the child-to-parent edge set $P_l$ as the map that assigns each $v \in V_l$ its cluster index.
These operations are illustrated on the left side of Figure \ref{fig:graphbuild}.

We use four pairs of affinity and threshold functions in this work, each meant to implement a human vision-inspired principle of perceptual grouping.
These principles are ordered by increasing ``strength,'' as they are able to group regions of the base tensor (and thus the input images) that are increasingly distinct in their visual features;
this is because the higher principles rely on increasingly strong \textit{physical assumptions} about the input.
The stronger principles P3 and P4 are inspired by the observation that infants initially group visual elements into objects based \textit{almost exclusively} on shared motion and surface cohesion, and only later use other features, such as color and ``good continuation'' -- perhaps after learning which visual cues best predict motion-in-concert \cite{Spelke1990}. 
For graph clustering, we use the Label Propagation algorithm because it does not make assumptions about the number of ``true'' clusters in its input and is fast enough to operate online during training of a PSGNet. 
However, any algorithm that meets these requirements could take its place.

\textbf{Implementation.}
The four types of affinity and thresholding function are defined as follows:

\textit{Principle P1: Feature Similarity.} Let $\mathbf{v} := A_l(v)$ denote the attribute vector associated with node $v$ at graph level $l$ (excluding its time-indexing and spatial centroid components.) Then the P1 affinity between two nodes $v,w$ is the reciprocal of their L2 distance, gated by a binary spatial window
\begin{equation}
        D^1(\vv, \ww) = \frac{\id(||c(v) - c(w)||_m < \dd)}{||\vv - \ww||_2},
\end{equation}
where $c(v)$ is the centroid of node $v$ in $(i,j)$ coordinates given by its spatial registration and $||\cdot||_m$ denotes Manhattan distance in the feature grid.
The P1 affinities are thresholded by the reciprocal of their local averages,
\begin{equation}
    \ep^1(D^1(v), D^1(w))= \min\left(\frac{1}{\overline{D}^1(v)}, \frac{1}{\overline{D}^1(w)}\right),
\end{equation}
where $D^1(v)$ denotes the set of nonzero affinities with node $v$ (i.e. affinities to nodes within its spatial window) and $\overline{D}^1(v)$ its mean.
Thus P1 produces binary edges $D^1(\vv, \ww) > \ep^1$ between nodes whose attribute L2 distance is less than the spatially local average, without learned parameters.

\textit{Principle P2: Statistical Co-occurrence Similarity.} This principle encodes the idea that if two nodes appear often in the same pairwise arrangement, it may be because they are part of an object that moves (or exists) as a coherent whole.
Thus it is a way of trying to infer motion-in-concert (the ground truth definition of an object that we use in this work) without actually observing motion, as when visual inputs are single frames.
This is implemented by making $D_{\phi_2}^2(\vv, \ww)$ inversely proportional to the reconstruction error of a Variational Autoencoder (VAE, \cite{kingma2013auto}) $H_{\phi_2}$, so that common node attribute pairs will tend to be reconstructed better (and have higher affinity) than rare pairs. Formally,
\begin{equation}
    \mathbf{e}_{vw} := |\vv - \ww|,
\end{equation}
\begin{equation}
    \hat{\mathbf{e}}_{vw} := H_{\phi_2}(\mathbf{e}_{vw}),
\end{equation}
\begin{equation}
    D^2(\vv, \ww) = \frac{1}{1 + \nu_2 \cdot ||\mathbf{e}_{vw} - \hat{\mathbf{e}}_{vw}||_2},
\end{equation}
where $\nu_2$ is a hyperparameter and both $\mathbf{e}_{vw}$ and $\hat{\mathbf{e}}_{vw}$ are vectors of the same dimension as $\vv$ and $\ww$.
This and all other VAEs in this work (see below) are trained with the beta-VAE loss \cite{higgins2017beta}, 
\begin{equation}
    \loss_{\text{VAE}} = ||\mathbf{e}_{vw} - \hat{\mathbf{e}}_{vw}||_2 + \beta \loss_{\text{KL}}(\hat{\mathbf{\mu}}_{vw}, \hat{\mathbf{\sigma}}_{vw}),
\end{equation}
where $\beta$ is a scale factor and $\loss_{\text{KL}}$ is the KL-divergence between the standard multivariate unit normal distribution and the normal distributions defined by the input-inferred $\mu$ and $\sigma$ vectors (VAE latent states.)
The $D^2$ affinities are thresholded at $0.5$ for all node pairs to produce binary edges.

\textit{Principle P3: Motion-driven Similarity.}
Parts of a scene that are \textit{observed} moving together should be grouped together, according to our physical definition of an object. 
This could be implemented by tracking nodes across frames (e.g. finding nearest neighbors in terms of nodes' attributes) and then defining co-moving nodes as those whose relative Eucldean distance (according to their $(x,y,z)$ attributes) changes less than some threshold amount. 
Because spatial and motion inference may have different amounts of observation noise for different nodes, though, we instead extend the VAE concept above to compute ``soft'' node affinities across and within frames.
Let $H^w_{\phi^w_3}$ and $H^a_{\phi^a_3}$ be two new VAE functions and $\nu^w_3$ and $\nu^a_3$ two new hyperparameters. 
Then, just as above, these define affinity functions $D^{3w}(\vv, \ww)$ and $D^{3a}(\vv, \mathbf{u})$ between nodes within each frame (``spatial'') and across adjacent frames (``temporal,'' i.e. $v \in V_l^t, u \in V_l^{t+1}$), respectively.
The only difference between the P2 and P3 affinities (other than having separate parameters) is that the latter are trained and evaluated \textit{only on nodes whose motion attributes}, $\Delta(v)$, \textit{are greater than a fixed threshold.}
This means that the P3 VAEs only learn which nodes commonly \textit{move} together, not merely \textit{appear} together.
Grouping \textit{via} the P3 edges (affinities thresholded at 0.5) then proceeds by clustering the entire spatiotemporal graph, $G_l = (\bigcup_t V^t_l, \bigcup_t (E^{3w}_t \cup E^{3a}_t))$.

\textit{Principle P4: Self-supervised Similarity from Motion.}
The final principle is based on the assumption that if two nodes are seen moving together, then nodes with similar attributes and relationships should be grouped together in future trials -- even if those nodes are not moving.
This implies a form of self-supervised learning, where the signal of motion-in-concert acts as an instructor for a separate affinity function that does not require motion for evaluation.
In principle, any supervision signal could be used -- including ``ground truth'' physical connections, if they were known -- but here we assume that motion-in-concert provides the most accurate signal of whether two visual elements are linked.
We therefore implement the P4 affinity function as a multilayer perceptron (MLP) on differences between node attribute vectors,
\begin{equation}
    D^4(\vv, \ww) = \sigma(Y_{\phi_4}(|\vv' - \ww'|)),
\end{equation}
where $\vv'$ is the attribute vector of node $v$ with its motion attribute $\Delta(v)$ removed, $Y_{\phi_4}$ is an MLP, and $\sigma(\cdot)$ is the standard sigmoid function, which compresses the output affinities into the range $(0,1)$.
The MLP weights are trained by the cross-entropy loss on the self-supervision signal from P3,
\begin{equation}
    \loss_{P4} = \sum_{v,w}\textbf{CE}(D^4(\vv, \ww), \id(P_3(v) = P_3(w))),
\end{equation}
where the ``ground truth'' is the indicator function on whether $v$ and $w$ have the same parent node, $P_3(v) = P_3(w)$, according to a round of P3 grouping from motion.
The binary P4 edges are obtained by thresholding the affinities at 0.5.

\textit{Label Propagation.}
To construct new parent edges, nodes are clustered according to within-layer edges from one of the four affinity functions using the standard Label Propagation (LP) algorithm \cite{labelprop} (Fig. \ref{fig:graphbuild} left, middle column.)
This algorithm takes as input only the edges $E_l$, the number of nodes at the current graph level $|V_l|$, and a parameter setting the number of iterations.
Each node is initialized to belong in its own cluster, giving a set of labels $[|V_l|]$.
Then for $q > 0$, iteration-$q$ labels are produced from iteration $q-1$-labels by assigning to each node $v \in V_l$ the most common stage-$q-1$ label among nodes connected to $v$ by an edge.
Ties are resolved randomly.
Let $P_l: V_l \rightarrow [m]$ denote the final converged label-prop assignment of nodes in $V$ to cluster identities, where $m$ is the number of clusters discovered.
Effectively, the new child-to-parent edges $P_l$ define \textit{input-dependent pooling kernels} for $V_l$ (Fig. \ref{fig:graphbuild} Left, right column.)
The final stage of building a new PSG level, \textit{Graph Vectorization}, uses these pooling kernels to aggregate statistics over the resulting partition of $V_l$, resulting in new nodes $V_{l+1}$ with $|V_{l+1}| = m$ (see below.)

\textbf{In this work.}
Two PSGNet architectures are used in this work: PSGNetS, which sees only static images (single-frame movies), and PSGNetM, which sees true movies and therefore can detect and learn from object motion. 
PSGNetS builds a three-level PSG by applying P1 grouping on the Level-0 nodes (ConvRNN features) to build Level-1 nodes, followed by P2 grouping on the Level-1 nodes to build Level-2 nodes (Fig. \ref{fig:graphbuild} right, red outline.)
PSGNetM builds a \textit{branched} four-level PSG by adding additional rounds of grouping to PSGNetS:
P3 grouping from motion-in-concert builds a set of \textit{spatiotemporal} Level-2 nodes (called Level-2M) from the Level-1 nodes;
and these self-supervise a round of P4 grouping on the original Level-2 nodes to build Level-3 nodes (Fig. \ref{fig:graphbuild} right.)
Other PSGNet architectures could build PSGs with different hierarchical structures by applying these (or new) grouping principles in a different pattern.
For instance, models that grouped from motion at different scales could, in principle, learn hierarchical decompositions of objects into their flexibly moving parts \cite{xu2019unsupervised}.

The VAEs used in P2 and P3 grouping are MLPs with hidden layer dimensions $(50, 10, 50)$; the $10$ dimensions in the latent state are considered as $5$ pairs of $(\mu, \sigma)$ parameters for independent normal distributions.
The $\beta$ scaling factor for each KL loss is $10$. 
The slope hyperparameters were set at $\nu_2 = 3.5$, $\nu^w_{3} = \nu^a_{3} = 10.0$ by grid search on a validation subset of \Play. 
The P4 MLP used in PSGNetM has hidden layer dimensions $(250, 250)$ and an output dimension of $1$.
In the P2 binary affinity function variant (BAff, Table \ref{tab:ablations} \cite{isola2015learning}), affinities are predicted by an MLP with hidden layer dimensions $(100, 100)$ and an output dimension of $1$.
All applications of LP proceed for $10$ iterations, which is enough to reach convergence (i.e., unchanging label identity with further iterations) for the majority of nodes.

\subsection{Graph Vectorization}
Given nodes $V_l$ and pooling kernels defined by $P_l$ from the previous stage, Graph Vectorization constructs a new level of nodes $V_{l+1}$ and their associated attributes $A_{l+1}: V_{l+1} \rightarrow \R^{C_{l+1}}$.
This also specifies a new spatial registration $R_{l+1}$, which maps each spatial index into the Level-0 nodes, $(i,j)$, to the index of its unique parent in $V_{l+1}$.
We call this process ``Vectorization'' because it encodes the group properties of a set of entities (here, lower-level graph nodes) as components of a single attribute vector.

New node attributes are created in two ways: 
first, through permutation-invariant aggregation functions of the lower-level nodes in each segment, $\textbf{Seg}[v] := P^{-1}_l(v)$ for $v \in V_{l+1}$;
and second, by applying MLPs to these aggregates.
The reason for computing the latter set of attributes is that aggregate statistics are unlikely to provide the direct encoding of scene properties (color, depth, normals, etc.) needed for Graph Rendering.
The use of MLPs allows these direct encodings to be learned as nonlinear functions of the aggregates, with parameters shared across nodes (much as standard convolution kernels share parameters across spatial positions.)

\textbf{Implementation.} 
For aggregation, we compute the mean of the first and second power of each lower-level node attribute as described in the main text. 
In addition to aggregating over the set $\Seg[v]$ for each node, we also compute aggregates over nine subsets of $\Seg[v]$ defined with respect to the spatial registration $R_{l+1}$:
its boundary, i.e. the set of nodes $w \in \Seg[v]$ whose registrations in the base tensor are adjacent to those of a node from a different segment $\Seg[v'], v' \neq v$;
its four ``quadrants,'' the subsets of nodes whose registered centroids are above and to the right, above and to the left, etc., of the centroid of $\Seg[v]$;
and the four boundaries of those quadrants.
Thus if nodes in $V_l$ have $C_l$ attribute components (including their spatial centroids), the new aggregates $A^{\textbf{agg}}_{l+1}(v)$ will have $2 \cdot 10 \cdot C_l$.

Two MLPs produce new attributes from these aggregates:  
A ``unary'' function $H^U_{l+1}(A^{\textbf{agg}}_{l+1}(v))$, which operates on each aggregate in $V_{l+1}$;
and a ``binary'' function $H^B_{l+1}(|A^{\textbf{agg}}_{l+1}(v) -  A^{\textbf{agg}}_{l+1}(w)|)$, which operates on pairs of aggregates.
(This latter operation is a type of graph convolution on the fully-connected graph of the Level-$l+1$ aggregates.)
The final set of new attributes $H^{\textbf{new}}(v)$ is given by 
\begin{equation}
    H^{\textbf{new}}_{l+1}(v) = H^U_{l+1}(v) + \frac{1}{|V_{l+1}|}\sum_{w \in V_{l+1}}H^B_{l+1}(v,w),
\end{equation}
that is, by summing the unary MLP outputs and the mean of the binary MLP outputs across node pairs.
The aggregate attributes and the learned attributes are concatenated to give the full attribute labeling maps, $A_{l+1}: v \mapsto A^{\textbf{agg}}_{l+1}(v) \oplus H^{\textbf{new}}_{l+1}(v)$.

\textbf{In this work.}
All MLPs used for Graph Vectorization have hidden layers with dimensions $(100, 100)$. 
The number of output dimensions is based on how many attributes are needed for rendering ($\QSR$ and $\QTR$). Each channel of an output $\QTR$ feature map requires $6$ components, and each of $6$ $\QSR$ constraints requires $4$ components. 
Because the full vectorization process would cause the number of node attributes to grow geometrically in the number of PSG levels, we compute only mean attributes and perform only constant texture rendering (requiring a single component per output channel) for Level-1 nodes;
in practice, these nodes have ``superpixel''-like spatial registrations with little feature variation across their aggregation domains and with similar shapes.

\section{Datasets, Baseline Models, and Evaluation Metrics}

\subsection{Datasets}
\textbf{Primitives.}
This dataset was generated in a Unity 2019 environment by randomly sampling 1-4 objects from a set of 13 primitive shapes (Cube, Sphere, Pyramid, Cylinder, Torus, Bowl, Dumbbell, Pentagonal Prism, Pipe, Cone, Octahedron, Ring, Triangular Prism) and placing them in a square room, applying a random force on one object towards another, and recording the objects' interaction for 64 frame-long trials.
The textures and colors of the background, lighting, and each object were randomly varied from trial to trial.
The training set has in total $180{,}000$ frames with 2 and 3 objects.
The test set has in total $4{,}000$ frames with 1-4 objects.
Each frame includes an image, a depth map, a surface normals map, a segmentation map of the foreground objects (used only for evaluation), and the camera intrinsic parameters (which are constant across each dataset.)
During training and evaluation of MONet, IODINE, and PSGNetS, individual frames across different trials are randomly shuffled, so object motion is not apparent from one frame to the next.

\textbf{Playroom}
This dataset was generated using the same Unity 2019 environment as \Prim\ with two differences: 
first, the floor and walls of the room were given naturalistic textures (including natural lighting through windows in the walls and ceiling);
and second, the objects were drawn from a much larger database of realistic models, which have naturalistic and complex textures rather than monochromatic ones. 
For each of $1000$ training trials or $100$ test trials, 1-3 objects were randomly drawn from the database and pushed to collide with one another.
Each trial lasts $200$ frames, giving a total of $200{,}000$ frames in the training set and $20{,}000$ in the testing set.
$100$ of the training trials were randomly removed to be used as a validation set.
When presented to PSGNetM and OP3, multi-frame clips from each trial were randomly selected and shuffled with clips from other trials, so object motion was apparent in a subset of training examples.
Both datasets will be made available upon request, and the Unity 2019 environment for generating these data (ThreeDWorld) will be available upon publication.

\textbf{Gibson.}
We used a subset of the Gibson 1.0 environment\footnote{http://buildingparser.stanford.edu/dataset.html}.
The subset is scanned inside buildings on the Stanford campus, and is subdivided into 6 areas.
We used area 1, 2, 3, 4, 6 for training, half of area 5 for validation, and the other half of area 5 for testing.
The training set has $50{,}903$ images and the validation and test set each have $8{,}976$ images, along with depth maps, surface normals maps (computed approximately from the depth maps), full-field instance segmentation labels, and the camera intrinsic parameters.

\subsection{Model and Training Implementations}
% \subsection{Model and Hyperparameter Details}

We implemented MONet, IODINE, and OP3 as described in the original papers and publicly available code \cite{monet, iodine, op3}.
We used the same convolutional encoder and decoder for all three models.
The encoder has 4 layers with (32, 32, 64, 64) channels each.
The decoder also has 4 layers with (32, 32, 32, 32) channels each.
The number of object slots $K$ is set to $7$ for all models on the synthetic datasets and $12$ for \textbf{Gibson}.
MONet uses a 5-block U-Net for attention.
It has in total $14.9$M parameters.
IODINE and OP3 use 5-step iterative inference \cite{iodine, op3}.
IODINE has $1.7$M parameters.
OP3 has $1.6$M parameters.
PSGNetS has $1$M parameters and PSGNetM $1.3$M. 

For Quickshift++, we used $1{,}000$ images from each training set to search for the best hyperparameters.
Table \ref{tab:q++_hps} shows the hyperparameters we found for each dataset.

\begin{table}[h]
    \setlength{\tabcolsep}{5pt}
    \centering
    \caption{Quickshift++ hyperparameters for each dataset.}
    %\vspace{-5pt}
    \begin{tabular}{lcc}
    \toprule
    Dataset & $k$ & $\beta$ \\
    \midrule
    Primitives & 20 & 0.9  \\
    Playroom & 20 & 0.95  \\
    Gibson & 80 & 0.95 \\
    \bottomrule
    \end{tabular}
    % \vspace{-15pt}
    \label{tab:q++_hps}
\end{table}

\textbf{Training.} We trained baseline models with the Adam optimizer \cite{kingma2016improved} with learning rate $0.0001$ and batch size $128$.
Gradients with norm greater than $5.0$ were clipped.
MONet was trained with 4 Titan-X GPUs.
IODINE and OP3 were each trained with 8 GPUs.
The training took between 48 and 96 hours to converge, depending on the model and dataset. 
PSGNetS and PSGNetM each were trained with batch size $4$ on a single Titan-X GPU for 24 hours using the Adam optimizer \cite{kingma2016improved} with learning rate $0.0002$. 
All models are based on CNN backbones, so can take images of any size as input;
however, training on images $>64\times64$ was computationally prohibitive for the baseline models, so we trained all models and report all evaluation metrics on images of this size.
PSGNets take significantly less time and resources to train than the baselines, so for visualization we trained and evaluated versions of PSGNetS and PSGNetM on $128\times128$ images.
This increases segmentation metric performance by 5-10\% (data not shown.)
Tensorflow \cite{abadi2016tensorflow} code for training and evaluating PSGNet models will be made available upon publication.

\subsection{Evaluation}
We use standard metrics for object detection and segmentation accuracy with minor modifications, due to the fact that all tested models output full-field scene decompositions rather than object proposals:
each pixel in an input image is assigned to exactly one segment.
The \textbf{mIoU} metric is the mean across ground truth foreground objects of the \textit{intersection over union} (IoU) between a predicted and ground truth segment mask (a.k.a. the Jaccard Index.) 
Because one predicted segment might overlap with multiple ground truth segments or \textit{vice versa}, for this and the other metrics we found the optimal one-to-one matching between predictions and ground truth through linear assignment, using $1.0 - \textbf{mIoU}$ as the matching cost. 
The \textbf{Recall} metric is the proportion of ground truth foreground objects in an image whose IoU with a predicted mask is $>0.50$.
\textbf{BoundF} is the standard F-measure (a.k.a. F1-score) on the ground truth and predicted boundary pixels of each segment, averaged across ground truth objects.
\textbf{ARI} (Adjusted Rand Index) is the Permutation Model-adjusted proportion of pixel pairs correctly assigned to the same or different segments according to the ground truth segmentation, as used in \cite{iodine} -- except that we do not restrict evaluation to foreground objects.
Linear assignment, ARI, and segment boundary extraction are implemented in Scikit-learn \cite{scikit-learn}.

In the main text we report metrics on each test set for single model checkpoints, chosen by where \textbf{Recall} on each validation set peaks (within 400 training epochs for the baselines and 5 training epochs for PSGNets.) 
Variation in PSGNet performance due to random weight initialization was <2\% across $5$ models.

\section{Comparing PSGNet to Baseline Models}

\begin{figure}[t]
\centering
    \includegraphics[width=1.0\textwidth]{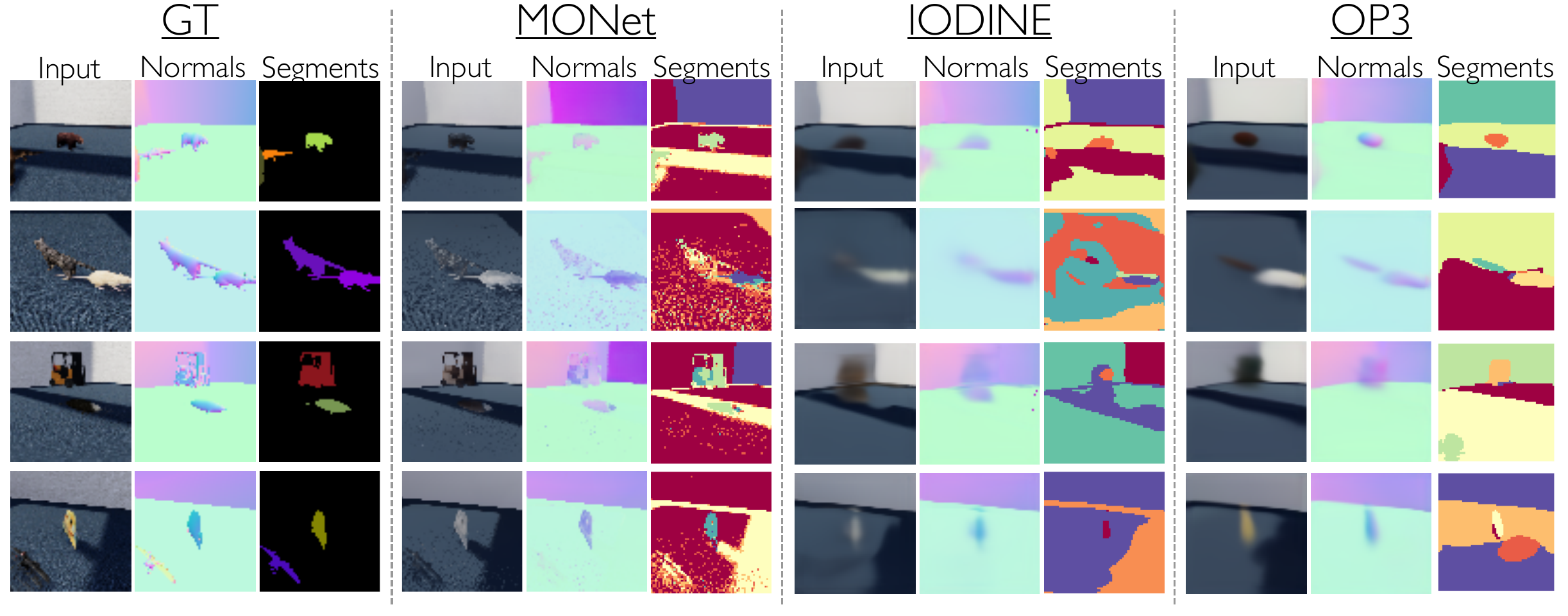}
\vspace{-15pt}
\caption{Examples (one per row) of the CNN-based models' RGB, normals, and segmentation predictions on \Play. (All models also receive supervision and make predictions in the depth channel, not shown here.) Unlike on \Prim, the baselines generally do not capture the shapes and textures of the objects in this dataset.}
\label{fig:baselines}
% \vspace{-15pt}
\end{figure}

\begin{figure}[t]
\centering
    \includegraphics[width=1.0\textwidth]{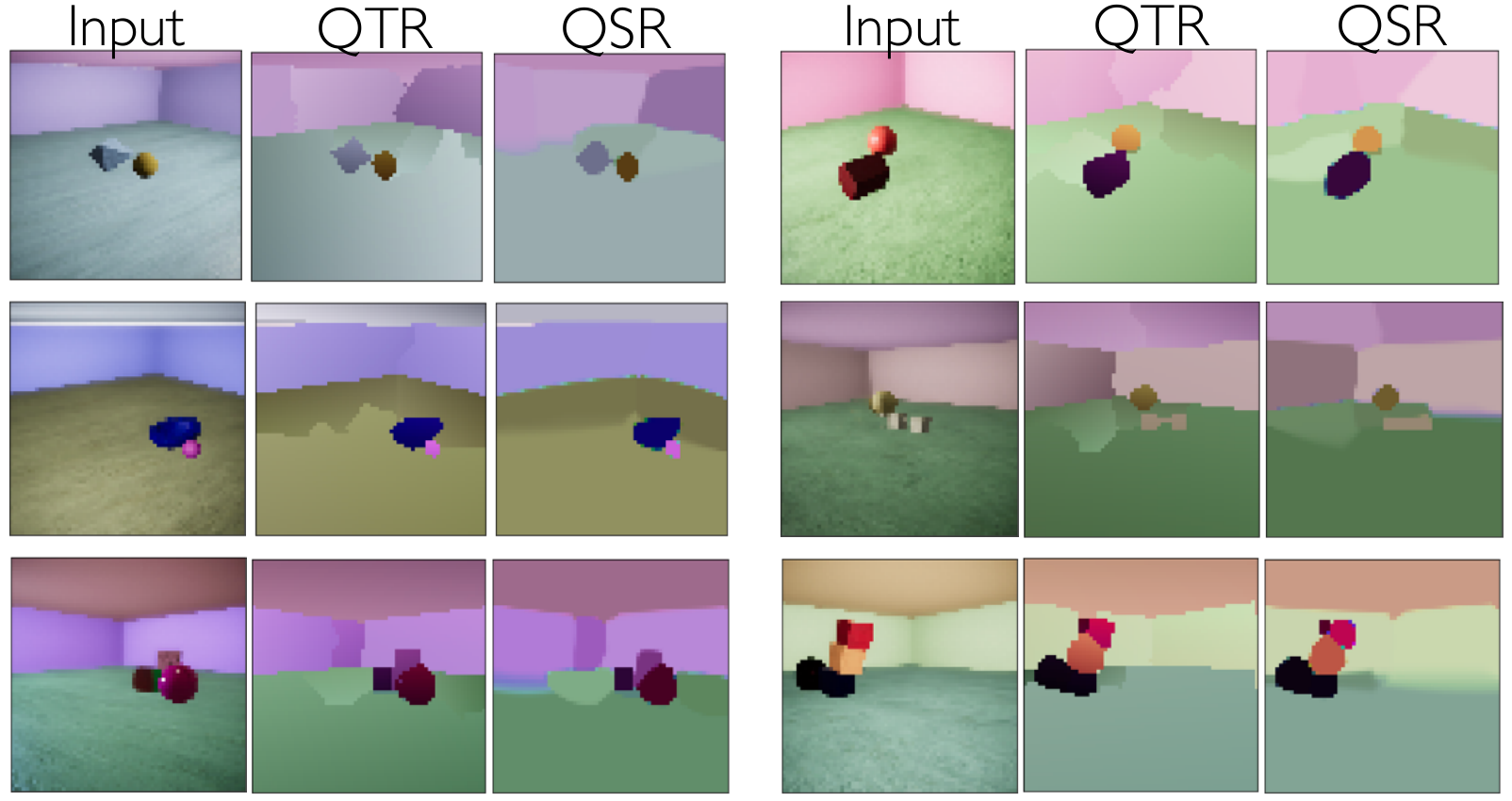}
\vspace{-15pt}
\caption{Six examples of PSGNetS top level (Level2) quadratic texture renderings of color (QTRs) and quadratic shape rendering (QSRs) on \Prim. Here the QSRs are filled in with the color of their associated nodes' color attributes, but they could be filled in with any other attribute. The QSRs are able to capture many of the simple silhouettes of objects in this dataset (compare to th QTRs, which are colored according to the Level2 unsupervised segmentations of each image.) Some silhouettes, such as that of the dumbbell (middle row, right column) are approximated by simpler shapes; adding additional quadratic constraints to the QSRs could produce more complex shapes.}
\label{fig:qsrs}
% \vspace{-15pt}
\end{figure}

\begin{figure}[t]
\centering
    \includegraphics[width=1.0\textwidth]{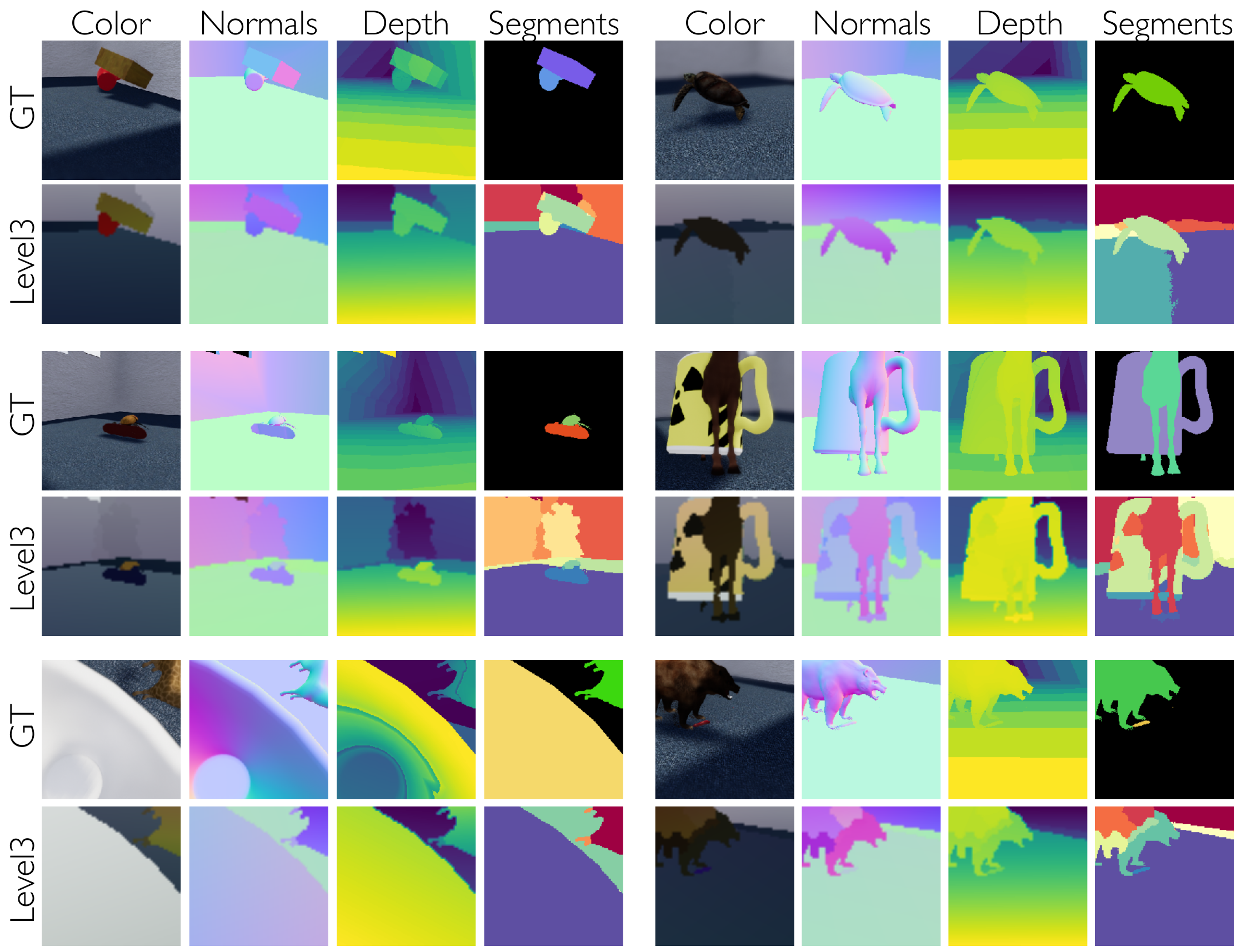}
\vspace{-15pt}
\caption{Six examples of PSGNetM renderings and segmentations on \Play. Only the top-level (Level3) quadratic texture renderings and spatial registration are shown, along with the ground truth for each input image. Note that the ground truth segmentations are not provided during training and only the color image is input to the model. The main failure mode of PSGNetM is undergrouping of large, static regions (e.g. first row, right column) or of objects with regions of very different textures (e.g. second and third rows, right column.)}
\label{fig:psgegs}
% \vspace{-15pt}
\end{figure}

Here we further describe the qualitative and quantitative differences between PSGNets and the CNN baselines as unsupervised scene decomposition methods.
At a high level, these methods produce such different results because (1) they make different assumptions about how scenes are structured and (2) they use different architectures (and therefore have different ``inductive biases'') to learn segmentations. 
The CNN-based models MONet \cite{monet}, IODINE \cite{iodine}, OP3 \cite{op3}, and related methods \cite{engelcke2019genesis} all assume that the visual appearance of a scene is generated by the combined appearances of up to $K$ latent factors. 
In essence, each model tries to infer the the parameters of the latent factors using a CNN encoder (or ``de-renderer'') and then reconstructs the scene with a decoder (``renderer'') that operates on each factor and combines their results into an output image.
These models therefore learn which scene components (i.e., segments of scenes with a certain 2D shape and appearance) are common across a dataset;
over the course of training, they get better at detecting and reconstructing these components.

In contrast, PSGNets do not learn \textit{via} the assumption that scenes can be decomposed into the appearance of several latent factors. 
They instead assume that \textit{pairs of visual elements are physically connected}, and therefore can be represented as parts of a single, higher-level entity.
This difference in assumptions is subtle but yields a dramatically different optimization problem for PSGNets, because they do not need to learn which object silhouettes (segmentation masks) or global appearances are common across a dataset;
the only need to learn which pairwise relationships appear frequently or persist through time (e.g. during object motion.)
Whereas silhouettes and global appearances vary widely even for the same object across different views and contexts, pairwise relationships (such as differences in color, texture, or surface normal vectors) are highly constrained by the physics of real, solid objects:
many are made of similar material and have smoothly changing shape across broad regions of their surfaces. 
In cases where objects have visually and geometrically distinct subregions, their concerted motion should reveal the ``ground truth'' of their underlying cohesion.

In addition to this difference in assumptions for learning, PSGNets have a major architectural difference with CNN-based methods:
they produce segments by \textit{spatially non-uniform grouping of scene elements} rather than fixed-size convolution kernels whose weights are shared across visual space.
It has previously been recognized that spatially uniform convolutions are poorly suited to produce sharp scene segmentations, motivating post-processing algorithms like conditional random fields for sharpening CNN outputs \cite{chen2017deeplab}.
At heart, this is because region borders and interiors tend to have different CNN feature activations (as spatially uniform receptive fields inevitably cross boundaries) even when they have the same visual appearance.
This problem is greater in the unsupervised setting, where there are no segmentation labels to indicate exactly where a boundary occurs.
Thus, we hypothesized that the perceptual grouping mechanisms of PSGNets would produce sharper boundaries and be less prone to overfitting common object shapes than the CNN-based methods.
This is consistent with the large previous body of work on \textit{unlearned} graph-based perceptual grouping and scene segmentation (e.g. \cite{shi2000normalized, arbelaez2010contour}). 
We extend the ideas behind hierarchical graph-based segmentation by learning image features and affinity functions according to physical properties of scenes.

\textbf{Qualitative comparison of model error patterns.}
Each model makes characteristic errors that can be explained by their respective architectures and loss functions.
MONet reconstructs the input image (and here predicts depth and normals feature maps) by producing one segmentation mask at a time from a U-Net CNN \citep{monet}, but imposes no further constraints on the structure of its outputs or latent states.
As a result its segmentations and predictions can have high spatial resolution and low pixel-wise reconstruction error even when inferring clearly non-physical "objects," such as segments split across disconnected regions of the image;
moreover, it learns to lump together disparate background regions because the expressive U-Net decoder can easily output single segment feature maps that adequately reconstruct predictable chunks of the scene (Fig. \ref{fig:baselines}, second column.)

IODINE (and its dynamics-modeling successor OP3) instead jointly infers all the parameters of all scene latent factors in parallel and "spatially broadcasts" them over an intermediate feature map at the same resolution as its output \citep{iodine}.
This encourages each predicted "object" to occupy a single, spatially contiguous region, but also appears to degrade information about its shape;
most foreground segment predictions resemble "blobs" that lack the distinguishing features of the shapes they represent (Fig. \ref{fig:baselines}, third and fourth columns.)

Finally, PSGNetS segments the scene into regions that are both contiguous and high-resolution, but sometimes oversegments -- that is, \textit{undergroups} during learnable graph pooling.
This is especially apparent for scene elements that are large (likely because learned affinities tend to fall off with distance) or contain strong shadows or reflections (because pairwise affinities may be low across a large change in appearance; see Figs. \ref{fig:qsrs} and \ref{fig:psgegs}.) 
Some of these errors are reduced in PSGNetM, which can group visually distinct regions that have been observed moving together. 
However, the addition of P3 and P4 in this model can lead to overgrouping when two scene elements with similar appearance are or have been seen moving together (see Fig. 2C in the main text.)
These two principles also do not provide learning signals about how to group scene elements that rarely or never move (such as floors and walls.)
In future work, we will explore additional grouping principles that can apply to these scene elements and can better distinguish moving objects even when they share similar appearance.

% \begin{table}[t]
%     \setlength{\tabcolsep}{5pt}
%     \centering\small
%     \caption{Training epochs before peak validation set object recall.}
%     %\vspace{-5pt}
%     \begin{tabular}{lcc}
%     \toprule
%     Model & Primitives & Playroom \\
%     \midrule
%     MONet & 200 & 38 \\
%     IODINE & 350 & 83 \\
%     OP3 & - & 24 \\
%     PSGNetS & 2.0 & 1.8 \\
%     PSGNetM & - & 1.0 \\
%     \bottomrule
%     \end{tabular}
%     \vspace{-15pt}
%     \label{tab:learning}
% \end{table}

\begin{table}[t]
    \setlength{\tabcolsep}{5pt}
    \centering\small
    \caption{Training epochs before peak validation set object recall.}
    %\vspace{-5pt}
    \begin{tabular}{lccccc}
    \toprule
    Dataset & MONet & IODINE & OP3 & PSGNetS & PSGNetM \\
    \midrule
    Primitives & 200 & 350 & - & 2.0 & - \\
    Playroom & 38 & 83 & 24 & 1.8 & 1.0 \\
    \bottomrule
    \end{tabular}
    \vspace{-15pt}
    \label{tab:learning}
\end{table}

\textbf{Learning efficiency and generalization.}
Beyond the gap between PSGNets and CNN baselines in segmentation performance, the substantially greater learning efficiency and generalization ability of PSGNets suggest that their bottom-up perceptual grouping architecture is better suited for scene decomposition than the latent factor inference architectures of the baselines.
On \Prim, where all models were able to decompose scenes reasonably well, peak object detection performance (\textbf{Recall}) on the validation set occurred after 200 and 350 training epochs for MONet and IODINE, respectively, but after only 2 epochs for PSGNetS (Table \ref{tab:learning}.)
The difference was less pronounced when training on \Play, where MONet, IODINE, and OP3 peaked after 38, 83, and 24 epochs -- compared to 1.8 and 1.0 for PSGNetS and PSGNetM;
however, the poor performance of the baselines on this dataset (<0.30 test set \textbf{Recall}) makes their learning efficiency a less important metric.
For the same reason, it is hard to assess the across-dataset generalization of the baselines, since they all underfit one of the two synthetic datasets used in this work.
Given that none of them achieved >0.10 \textbf{Recall} when tested on the dataset they were not trained on, we consider their generalization ability limited.
In contrast, both \Prim-trained PSGNetS and \Play-trained PSGNetM achieved >50\% of their within-distribution test performance on their converse datasets, suggesting that a significant portion of their learned perceptual grouping applies generically to objects that have never been seen before.
This is our expectation for algorithms that purport to learn ``what objects are'' without supervision, rather than learning to detect a specific subset of objects they have seen many times.
Future work will explore what constitutes an optimal set of training data for the PSGNet models.

\textbf{Assessing the dependence of baselines on geometric feature maps.}
Finally, because the CNN baselines were all developed to solve an autoencoding (or future prediction) problem on RGB input only, we wondered whether the task used in this work -- reconstruction of supervising RGB, depth, and normals maps -- might have have stressed these models in unintended ways (even though they were given strictly more information about input scenes during training.)
Surprisingly, neither giving depth and normal maps as \textit{inputs} (in addition to the RGB image) nor reconstructing RGB alone substantially changed the performance of MONet or IODINE on \Prim:
both autoencoding variants of both models yielded performance within 10\% of what they achieved on the original RGB to RGB, depth, and normals task.
It is unclear why the baseline methods do not learn to decompose scenes better when given more information about their geometric structure (cf. the NoDN PSGNetM ablation.)
One possible explanation is that color cues alone are highly predictive of object boundaries in \Prim, whereas the depth and normals channels only rarely indicate boundaries that are not easily detected in the color channels.

\section{Visualizing PSGNet Outputs}

Here we provide additional examples of components of predicted PSGs. 
Figure \ref{fig:qsrs} compares the PSGNetS top-level (Level-2) color QTRs to the top-level QSRs on images from \Prim. 
The former are colored by quadratic ``painting by numbers'' in the Level-2 spatial registration inferred for each image, so better reconstruct both high-resolution details and smooth changes in color across large regions.
The latter are shapes ``drawn'' by intersecting $6$ predicted quadratic constraints per node, colored by each node's RGB attribute. 

Figure \ref{fig:psgegs} shows PSGNetM top-level (Level-3) QTRs for color, depth, and normals, as well as the top-level spatial registration, for single images in \Play. 
Despite the wide variety of object sizes, shapes, and textures in this dataset, PSGNetM is largely able to segment and reconstruct the attributes of most.
Undergrouping (oversegmentation) failures tend to occur when objects have sharp internal changes in color. 
Interestingly, human infants seem not to perceptually group visual elements by color in their first few months of life, relying instead on surface layout and motion \cite{Spelke1990};
it is possible that PSGNetM could better learn to group parts of these complex objects by explicitly ignoring the color and texture attributes of their nodes, which is feasible because of their direct and disentangled encoding.

\end{document}